%% file: LFM_survey.tex
\documentclass[10pt,journal,compsoc]{IEEEtran}
\usepackage{color}
\usepackage{caption}
\usepackage{times}
\usepackage{helvet}
\usepackage{booktabs} 

\usepackage{longtable}
\usepackage{array}
\usepackage{multirow}

\usepackage{courier}
\usepackage{amssymb}
\usepackage{amsthm}
\usepackage{bbm}
\usepackage{graphicx}
\usepackage{array}
\usepackage{booktabs}
\usepackage{amsopn}
\usepackage{amsmath,bm}
\usepackage{algorithm}
\usepackage{algorithmic}
\usepackage{enumerate}
\usepackage{color}

\usepackage{multirow}
\usepackage{bbding}
\usepackage{threeparttable}
\usepackage{threeparttable} 
\usepackage{dsfont} 
\usepackage{makecell} 
\usepackage{hyperref}
\usepackage{bm}
\usepackage{url}            
\usepackage{tabularx}
\newcommand{\eg}{\emph{e.g.}}

\newcommand{\ie}{\emph{i.e.}}

\begin{document}

\title{Learning from models beyond fine-tuning}

\author{
Hongling Zheng, Li Shen, Anke Tang, Yong Luo, Han Hu, Bo Du, Yonggang Wen, Dacheng Tao 
  \IEEEcompsocitemizethanks{\IEEEcompsocthanksitem Hongling Zheng, Anke Tang, Yong Luo, and Bo Du are with the School of Computer Science, School of Artificial Intelligence, National Engineering Research Center for Multimedia Software and Hubei Key Laboratory of Multimedia and Network Communication Engineering, Wuhan University, China. E-mail: \{hlzheng, anketang, dubo\}@whu.edu.cn, yluo180@gmail.com
  \IEEEcompsocthanksitem  Li Shen is with the School of Cyber Science and Technology, Shenzhen Campus of Sun Yat-sen University and JD Explore Academy, China. E-mail: mathshenli@gmail.com
 \IEEEcompsocthanksitem  Han Hu is with the Beijing Institute of Technology, China. E-mail: hhu@bit.edu.cn
  \IEEEcompsocthanksitem  Yonggang Wen and Dacheng Tao are with the Nanyang Technological University, Singapore. E-mail: ygwen@ntu.edu.sg, dacheng.tao@gmail.com
}
}

\markboth{Journal of \LaTeX\ Class Files,~Vol.~14, No.~8, October~2023}%
{Shell \MakeLowercase{\textit{et al.}}: Bare Demo of IEEEtran.cls for Computer Society Journals}

\IEEEtitleabstractindextext{%
  \begin{abstract}
    \input{section/abstract}
  \end{abstract}

  }

\maketitle

\IEEEdisplaynontitleabstractindextext

\IEEEpeerreviewmaketitle

\input{section/introduction}
\input{section/model_tuning}
\input{section/distillation}

\input{section/model_reuse}

\input{section/meta_learning}
\input{section/model_editing}
\input{section/challenges}
\input{section/conclusion}





\ifCLASSOPTIONcaptionsoff
  \newpage
\fi

\renewcommand\refname{References}

{\small
  \bibliographystyle{unsrt2authabbrvpp}
  \bibliography{LFM_survey}
}

\end{document}

%% file: section/abstract.tex
Foundation models (FM) have demonstrated remarkable performance across a wide range of tasks (especially in the fields of natural language processing and computer vision), primarily attributed to their ability to comprehend instructions and access extensive, high-quality data. This not only showcases their current effectiveness but also sets a promising trajectory towards the development of artificial general intelligence.
Unfortunately, due to multiple constraints, the raw data of the model used for large model training are often inaccessible, so the use of end-to-end models for downstream tasks has become a new research trend, which we call \textit{~\textbf{Learn From Model (LFM)}} in this article. LFM focuses on the research, modification, and design of FM based on the model interface, so as to better understand the model structure and weights (in a black box environment), and to generalize the model to downstream tasks. The study of LFM techniques can be broadly categorized into five major areas: model tuning, model distillation, model reuse, meta learning and model editing. Each category encompasses a repertoire of methods and strategies that aim to enhance the capabilities and performance of FM.
This paper gives a comprehensive review of the current methods based on FM from the perspective of LFM, in order to help readers better understand the current research status and ideas. To conclude, we summarize the survey by highlighting several critical areas for future exploration and addressing open issues that require further attention from the research community. 
The relevant papers we investigated in this article can be accessed at \url{https://github.com/ruthless-man/Awesome-Learn-from-Model}.

%% file: section/introduction.tex
\IEEEraisesectionheading{\section{Introduction}\label{sec:introduction}}

The rapid advancement of algorithms and computing power has sparked significant development and interest in large-scale pre-training models across both industry and academia.
These models, such as GPT-3~\cite{brown2020language}, LLAMA~\cite{touvron2023llama}, and Imagen~\cite{saharia2022photorealistic}, leverage the power of over-parameterized transformers to effectively model natural language in a variety of ways. This infrastructure enables these models to handle large-scale language and vision tasks and exhibit impressive performance across a wide spectrum of downstream applications. The continuous growth and refinement of FM indicate a promising future for natural language processing and related fields.
\begin{figure*}[ht]
  \centering
  \includegraphics[width=1\linewidth]{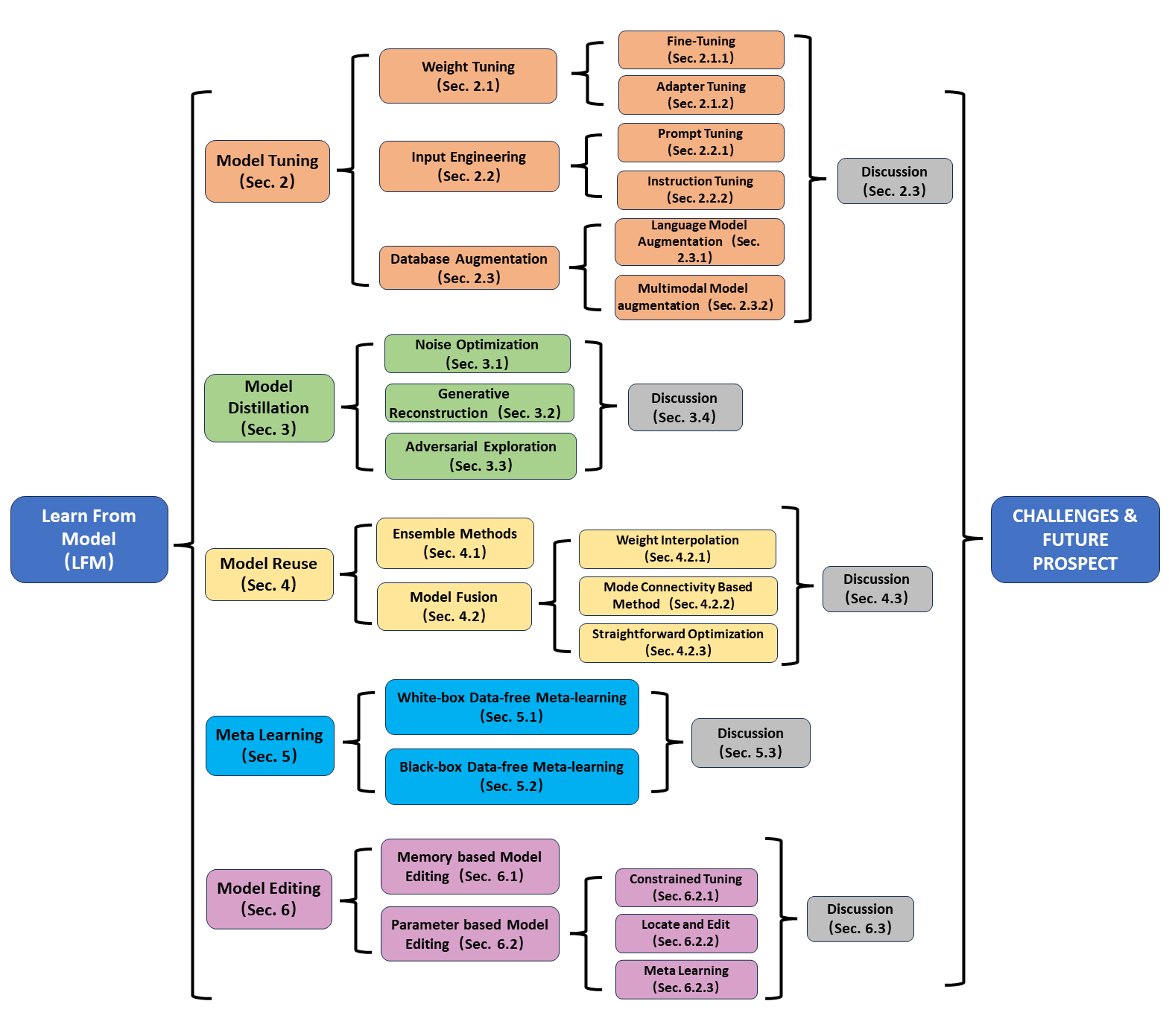}
  \caption{The structural taxonomy for LFM. The survey is organized according to the hierarchical structure.}
  \label{fig:LFM}
\end{figure*}

\begin{figure*}[ht]
  \centering
  \includegraphics[width=0.98\linewidth]{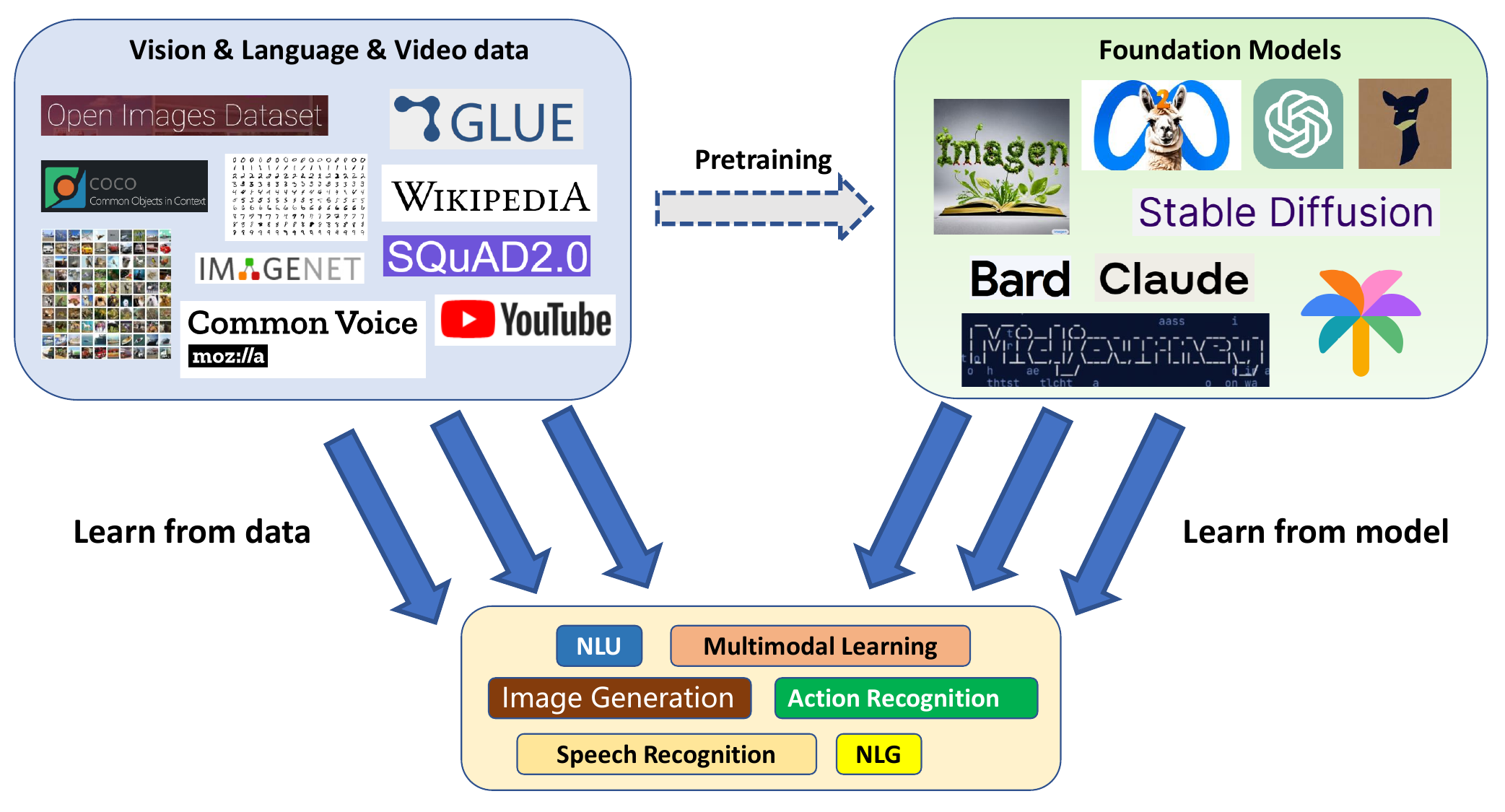}
  \caption{
    \textbf{Two fundamental approaches in machine learning}.
    The ``learn from data" and ``learn from model" are two fundamental approaches that play a vital role in training models and improving their performance. Both approaches have their strengths and applications.
    \textbf{LFD}: The "learn from data" approach involves training models by using large amounts of labeled or unlabeled data. This approach relies on the premise that patterns and relationships within the data can be learned by the model to make accurate predictions.
    \textbf{LDM}: The "learn from model" approach involves leveraging the knowledge and insights gained from existing FM to improve model performance. Rather than starting from scratch with raw data, this approach utilizes FM as a foundation and builds upon them.
  }
  \label{fig:overview}
\end{figure*}

Standing on the shoulders of giants, research based on FM are demonstrating unprecedented vigor and innovation in the current landscape~\cite{zamfirescu2023johnny,alberts2023large,ghosal2023text,jiang2020can,gou2021knowledge}. This gives rise to a burgeoning research paradigm, which we term as \textit{\textbf{Learn From Model (LFM)}} in this paper. As shown in Figure~\ref{fig:overview}, LFM refers to the study of foundation models to understand the model's behavior, strengths, and possible shortcomings. This kind of research can help us to better optimize the performance of the model, find and fix the problems of the model, and ultimately put the model into production and work better.

There are numerous reasons to learn from the model itself rather than from the data used to train the model. From a data privacy perspective, large models are typically trained on vast amounts of data, which may contain sensitive information such as personal identity information and private communications. Directly studying this data could infringe on user privacy~\cite{liu2023summary,wang2023model}. However, by studying the model itself, we can avoid direct contact with this sensitive data. At the same time, the data used to train large models involve commercial interests. These data, as a crucial part of the competitive advantage of the model owners, are usually not disclosed~\cite{carlini2021extracting}. Therefore, ordinary users can usually only access the API input interface and the final output of the model, which further promotes the motivation to learn from the model itself. From the perspective of model generalization ability, by studying the model itself, we can better understand the model generalization ability, that is, how the model handles new data it has not seen during training~\cite{anil2022exploring,wang2022language,zhong2022improving}. Both approaches, learn from data and learn from model, have their strengths and applications~\cite{misra2023reprogramming}. Learn from data is fundamental in situations where abundant labeled or unlabeled data is available, while learn from model is useful when existing models can provide valuable insights, speed up training, or enhance prediction accuracy. These approaches often complement each other in practice, enabling researchers and practitioners to build innovative and high-performing machine learning solutions.

In general, there are several advantages to learn from model: From a data perspective, due to factors such as data privacy and storage costs, it is difficult for smes and even individual users to obtain data to train high-quality models. The existing basic model highly compresses the characteristic information of large-scale raw data, and the requirement of raw data can be reduced as much as possible based on LFM paradigm. From a computing standpoint, as the parameters of the base model scale from 175B parameters of GPT-3 to 1.8T parameters of GPT-4, the computational power requirement of the retraining model increases exponentially ~\cite{dupuis2022heuristic}. The paradigm of the base model combined with downstream task tuning has become an alternative, with significant computational cost savings. From the perspective of knowledge representation, the Emergent Abilities~\cite{wei2022emergent} displayed by the large model are able to show extraordinary performance in general purpose tasks, and the mechanism behind this is difficult to obtain using the combination of small models. At the same time, the self-monitoring capability of the basic model can provide impetus and theoretical basis for downstream applications. Finally, from the perspective of the industry, FM APIs with carefully adjusted parameters have emerged one after another, and SOTA performance on new tasks based on high-quality FM has become a recognized paradigm. With the advent of the large model era, we can foresee that LFM will replace the traditional paradigm of learning from data and become a new way of thinking.

This paper categorizes the paradigms of FM according to their application scenarios, including model tuning, model distillation, model fusion, model reuse, meta learning and model editing. In the last part of the body of the survey, we analyze and provide a forecast for the future applications of FM.

The main job of model tuning~\cite{kuhn2013over} is to design the FM (modify internal parameters, add additional components) to achieve efficient performance on downstream tasks. This approach allows fast adaptation of the model to new tasks and avoids the large computational cost of training the model from scratch. The above techniques are all examples of LFM. Model distillation~\cite{gou2021knowledge} transfers knowledge from a large model (teacher model) to a small model (student model) that can operate in a resource-constrained environment while maintaining similar performance to the large model. Model Reuse~\cite{jiang2023llm} learns strengths from models, combining predictions from multiple models to improve overall performance. Meta-learning~\cite{hospedales2021meta}, also known as "learning to learn from model," as a way to design models so that they quickly adapt to new tasks, is similarly brought under the scope of LFM. With the rapid change of the real world, the knowledge inside the FM inevitably has the problem of insufficient accuracy. Model editing~\cite{mitchell2021fast} solves the defect of backward model knowledge by directly adjusting the model behavior, avoiding the huge resource consumption caused by retraining the model.

Our contributions can be summarized into these folds:
\begin{itemize}
  \item Based on a deep understanding of current research, we first summarize and propose the concept of LFM, which aims to outline research based on FM, liberating mindsets from traditional paradigm of learning from data.
  \item Compared with existing surveys, we provide a more systematic survey of LFM. Our survey includes specific classifications and cutting-edge analysis of LFM methods, as well as corresponding application trends, providing a more comprehensive overview of the field.
  \item Our study serves as a stepping stone for the scientific community, shedding light on the diverse opportunities and challenges that lie ahead in the pursuit of learning from large pre-trained models. We hope that our work will foster a deeper understanding of the LFM paradigm, catalyzing breakthroughs that will ultimately benefit both industry and academia in the years to come.
\end{itemize}

The whole article is structured as follows: We first introduced the definition and related classification of LFM technology in Section~\ref{sec:introduction}, described the model tuning in Section~\ref{sec:model tuning}, reviewed the model distillation technology in section~\ref{sec:model distillation}, reviewed the model reuse in section~\ref{sec:model reuse}, introduced the application of meta-learning in the LFM paradigm in section~\ref{sec:meta-learning}, and introduced the model editing in sections~\ref{sec:model editing} The future direction of LFM is discussed in Section~\ref{sec:challenges}. Finally, the thesis is summarized in Section~\ref{sec:conclusion}.

%% file: section/model_tuning.tex
\section{Model Tuning}
\label{sec:model tuning}





\begin{table*}[htb]
  \centering
  \begin{tabularx}{\textwidth}{p{0.3\linewidth} p{0.5\linewidth} p{0.2\linewidth}}
    \toprule
    \textbf{Literature}                                                        & \textbf{Summary}                                                                                                                                                                                     & \textbf{Classification} \\
    \midrule
    Big Self-Supervised Models ~\cite{NEURIPS2020_fcbc95cc}                    & Semi-supervised learning for large visual datasets.                                                                                                                                                  & Fine-Tuning             \\
    AIFT~\cite{Zhou_2017_CVPR}                                                 & Try to integrate active learning and transfer learning into one framework.                                                                                                                           & Fine-Tuning             \\
    GPT-1~\cite{radford2018improving}                                          & A generative model that is pre-trained on an unsupervised corpus and fine-tuned on a supervised dataset.                                                                                             & Fine-Tuning             \\
    R4F~\cite{aghajanyan2020better}                                            & An approach based on trust domain theory is proposed to replace adversarial targets with parametric noise, thereby reducing representation changes during fine-tuning without affecting performance. & Fine-Tuning             \\

    Lightweight Adapter Tuning~\cite{le2021lightweight}                        & Adapter adjustment for multilingual neural machine translation.                                                                                                                                      & Adapter Tuning          \\

    MHR~\cite{caccia2023multihead}                                             & New methods such as Multi-Head Routing (MHR) and MHR-$\mu$ are introduced to enhance presentation capabilities and optimize multitasking adaptation.                                                 & Adapter Tuning          \\

    ADAMIX~\cite{wang2022adamix}                                               & A parametric efficient method is proposed to increase adapter capacity in large pre-trained language models.                                                                                         & Adapter Tuning          \\

    LLM-Adapters~\cite{hu2023llm}                                              & Various adapters can be integrated into the LLM and these adapter-based LLM PEFT methods can be executed for different tasks.                                                                        & Adapter Tuning          \\

    Black Box Adversarial Prompting for Foundation Models~\cite{maus2023black} & A black box framework for generating adversarial prompts for unstructured images and text generation is developed.                                                                                   & Prompt Tuning           \\

    Prefix-Tuning~\cite{li2021prefixtuning}                                    & A black box framework for generating adversarial prompts for unstructured images and text generation is developed.                                                                                   & Prompt Tuning           \\

    Progressive Prompts~\cite{razdaibiedina2023progressive}                    & A simple and effective method for continuous language model learning.                                                                                                                                & Prompt Tuning           \\

    RLPrompt~\cite{deng2022rlprompt}                                           & Optimizing Discrete Text Prompts with Reinforcement Learning.                                                                                                                                        & Prompt Tuning           \\


    GRAM~\cite{li2023gradientregulated}                                        & A novel gradient-regulated meta-cue learning framework is introduced by using only unlabeled image-text pre-training data.                                                                           & Prompt Tuning           \\


    BBTv2~\cite{sun-etal-2022-bbtv2}                                           & The gradient-free adaptive of pre-trained models (PTMs) is enhanced by optimizing continuous prompts without accessing model parameters.                                                             & Prompt Tuning           \\

    MULTIINSTRUCT~\cite{xu2023multiinstruct}                                   & The first multimodal instruction tuning baseline dataset.                                                                                                                                            & Instruction Tuning      \\

    LINGUIST~\cite{rosenbaum2022linguist}                                      & Instruction fine-tuning of large-scale seq2seq models to control output generated from multilingual intents and slot markup data.                                                                    & Instruction Tuning      \\

    LLaVA~\cite{liu2023visual}                                                 & End-to-end trained large multimodal model that connects a vision encoder and LLM for general-purpose visual and language understanding.                                                              & Instruction Tuning      \\

    GPT4RoI~\cite{zhang2023gpt4roi}                                            & Visual features extracted by the spatial instruction and the language embedding are input to LLM, and trained on the data in instruction tuning format.                                              & Instruction Tuning.     \\

    \bottomrule
  \end{tabularx}
  \caption{Model Tuning Literature}
  \label{table:model-tuning}
\end{table*}

As one of the paradigms of LFM, the main goal of model tuning is to design the parameters of the FM for migration to downstream tasks. Compared with retraining model based on data for new tasks (learn from data), the idea of model tuning based on transfer learning reduces the hidden dangers of insufficient data sets and high training costs, and makes use of the common sense knowledge stored by the FM itself to provide convenience for parameter initialization for new tasks. The success of model tuning requires an in-depth study of the internal structure and dynamics of the pretrained model, understanding how the FM encode input data and identifying specific components that strongly influence the predicted results. On the other hand, according to the modification degree and modification position of model parameters, model tuning can be further divided into weight tuning, input engineering and database augmentation. Weight tuning is based on the internal parameters of FM, input engineering designs better suggestions from the input level, and database augmentation updates model knowledge based on external databases to ensure model accuracy.

\subsection{Weight Tuning}
\subsubsection{Fine-Tuning}

With the dramatic changes in model structure in recent years, methods such as fine-tuning have gradually replaced supervised learning~\cite{liu2023pre} as the main standard. Fine-tuning is a transfer learning technique for using the knowledge of pre-trained neural networks to solve new, relevant tasks. In the field of natural language processing (NLP), fine-tuning is often used to adjust large pre-trained language models (e.g. BERT, GPT, etc.) to specific tasks such as text classification, sentiment analysis, named entity recognition, etc. This approach can significantly reduce training time and improve the model's performance on new tasks~\cite{lester2021power}. Here we define the fine-tuning for updating all parameters of the model. By utilizing the knowledge of the pre-trained model, fine-tuning can significantly improve the model's ability to generalize on new tasks.

The idea of pre-training large models was well established before deep neural network modeling became the paradigm~\cite{min2021recent}. Nima~\cite{tajbakhsh2016convolutional} in 2016 investigated the effects of fine-tuning on the effectiveness of deep convolutional neural networks in the context of medical image analysis, demonstrating the robustness of layered fine-tuning. GPT-1~\cite{radford2018improving} first introduced fine-tuning ideas into the design of large language models, proving the effectiveness of this approach in a wide range of benchmarks for natural language understanding. BitFit~\cite{DBLP:journals/corr/abs-2106-10199} optimizes the bias parameters in pre-trained transformers, effectively improving performance on all evaluated GLUE tasks. This approach streamlines deployment and enhances hardware efficiency by leveraging pre-trained weights for the majority of computations, only modifying a minor fraction during inference. WiSE-FT~\cite{wortsman2022robust} improves the robustness of fine-tuned pre-trained models by ensembling the weights of the zero-shot and fine-tuned models. This method not only maintains high accuracy on a given target distribution, but also significantly boosts the model's performance under distribution shifts. 

Although there have been improvements and studies on traditional fine-tuning model methods\cite{aghajanyan2020better,kumar2022fine,ruiz2023dreambooth}, there are still inevitable major flaws in principle. To be specific, fine-tuning essentially requires that we can fully explore and master the internal parameters and structure of the model. On this basis, reverse gradient propagation is carried out to adjust a large number of model parameters, reuse the model, and generalize to a new task. As a result, there are several limitations in a real LFM scenario: \textbf{Computational Resource Requirement}: Full-parameter fine-tuning requires substantial computational resources because it involves optimizing all the parameters of the model. This could be problematic in a small-scale or resource-constrained environment~\cite{chen2020deep}. There is a notable risk of \textbf{Overfitting Risk}: In scenarios where the amount of training data is limited, full-parameter fine-tuning can lead to overfitting, where the model may perform well on the training data but not generalize effectively to new, invisible data~\cite{sun2022singular}. Another critical issue is \textbf{Catastrophic Forgetting}: Fine-tuning may cause the model to forget the pretraining knowledge it initially learned when it is trained on a new task. This phenomenon is also known as "catastrophic forgetting"~\cite{hayes2020remind,razdaibiedina2023progressive}. Concerns about \textbf{Model Stability Issues} are also worthy of attention: Fine-tuning could disrupt the stability of the model. The patterns learned by the pretrained model on a large corpus may be distorted during the fine-tuning process, leading to a decrease in the quality of the model's generated outputs~\cite{fu2023effectiveness}.



\subsubsection{Adapter Tuning}

Adapter tuning maintains most of the model parameters fixed and inserts new trainable parameters (referred to as "adapters") in between the internal layers of the model, specifically for the purpose of fine-tuning on specific tasks. This strategy was proposed by Houlsby~\cite{houlsby2019parameter} in 2019, with the aim of enhancing the fine-tuning performance of models while reducing computational costs and the risk of overfitting.
\begin{figure}[ht]
  \centering
  \includegraphics[width=1\columnwidth]{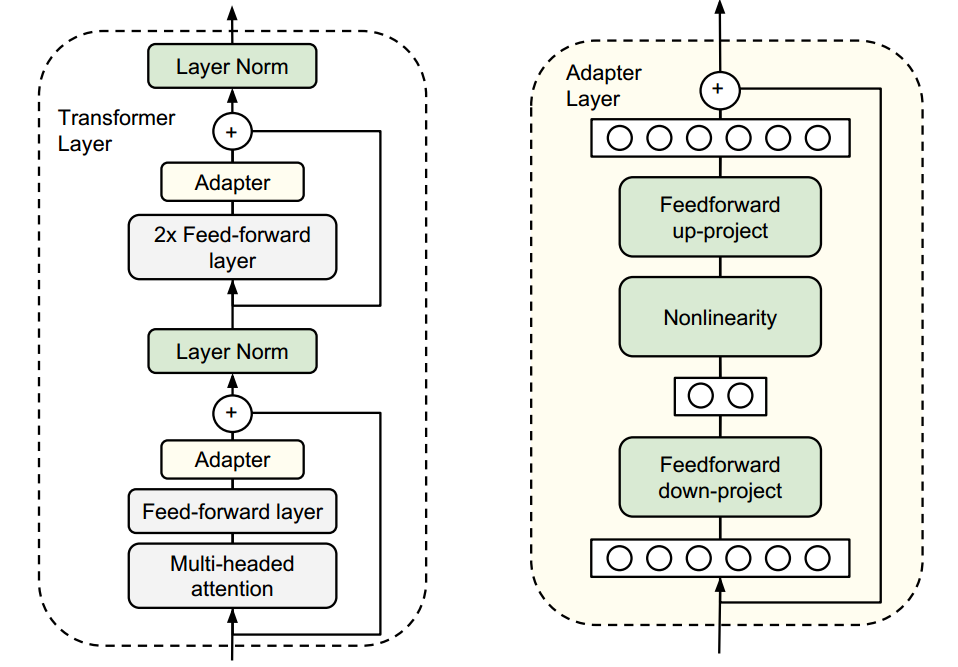}
  \caption{The basic structure of Adapter Tuning~\cite{houlsby2019parameter}.}
  \label{fig:Adapter Tuning}
\end{figure}
Adapters are inserted layer by layer to make the least modification to the original model structure and don not require retraining a large number of parameters, hence significantly reducing the computational cost in the fine-tuning process.
Specifically, one adapter module contains a down-projection and an up-projection. For an input feature h, a down-projection projects the input to a r-dimensional space with a parameter matrix $W_d$, after which a nonlinear function f is applied. Then the up-projection $W_u$ maps the r-dimensional representation back to d-dimensional space. Added with a residual connection, the complete computation could be written as: $$\hat{h}=f(h*W_d)*W_u+h.$$ Adapters can be trained for multiple tasks and then plugged into the pre-trained model to perform new tasks~\cite{ding2023parameter}, as shown in Figure~\ref{fig:Adapter Tuning}.

Based on the Adapter architecture, AdapterFusion~\cite{pfeiffer2020adapterfusion} realizes the maximum task migration between multiple adapter modules by separating the two stages of knowledge extraction and knowledge combination. He~\cite{he2021effectiveness} provides an in-depth study on the effectiveness of adapter-based tuning. Not only does adapter-based tuning mitigate forgetting issues by maintaining closer adherence to the initial pretrained
language model's representations, it also demonstrates superior performance on low-resource and cross-lingual tasks, exhibiting greater robustness to overfitting and changes in learning rates compared to standard fine-tuning.
Although an adapter works with much fewer tunable parameters than vanilla fine-tuning, some work attempts a more rigorous saving strategy by introducing inductive biases into the structure of the adapter layer. Rabeeh~\cite{NEURIPS2021_081be9fd} inserts the task-specific weight matrix into the FM's weight, which is efficiently calculated as the sum of the Kronecker product between the shared slow "weights" and the fast "first-ranked matrices defined for each compter layer. 
In multilingual speech translation (ST) tasks, Le~\cite{le2021lightweight} effectively integrated adapter modules for further specialisation of a fully-trained multilingual ST model for each language pair. 
AdaMix~\cite{wang2022adamix} introduces a parameter-efficient fine-tuning (PEFT) approach for large pre-trained language models. This method enhances performance on downstream tasks by leveraging a mixture of adaptation modules. Despite tuning only 0.1-0.2\% of parameters, AdaMix outperforms full model fine-tuning and existing PEFT methods.

Adapter tuning maintains performance comparable to full-parameter fine-tuning while reducing computational cost. Additionally, due to the fewer parameters of adapters, the risk of overfitting is also relatively reduced. Another significant advantage of adapters is that, as they are separately added to the model, they can be easily shared or transferred among different tasks, enabling multi-task learning or transfer learning with the model~\cite{chen2023exploring}. On the other hand, Adapter-based approaches require adding relevant parameters to the pre-trained model for downstream tasks. Although they improve the training efficiency of the model, they also lead to the problem of inference delay. When the model is deployed in real applications, the speed drop will be very noticeable~\cite{vander2023using}.

\subsection{Input Engineering}
\subsubsection{Prompt Tuning}

Downstream fine-tuning induces a high-cost burden by modifying all parameters. To alleviate the tuning burden and maintain the performance, researchers froze large pre-trained models and try to optimize the prompts. Brown~\cite{brown2020language} was the first to demonstrate the remarkable ability of prompt engineering to adapt large pre-trained models to downstream tasks. Comprehensive results~\cite{gu-etal-2022-ppt,lester-etal-2021-power} also confirm the optimal prompts could induce comparable ability with fine-tuning models.
In this part, we describe how to learn better prompts from the fixed FM. Concretely, we separate this application into two scenarios: 1) White-Box Prompt Tuning (i.e., having access to gradients and parameters); 2) Black-Box Prompt Tuning (i.e., having no access to gradients and parameters).

\textbf{White-Box Prompt Tuning:}
White-Boxing tuning built on emerging considerable open access models in early stages of FM development. 
Li~\cite{li-liang-2021-prefix} proposed prefix-tuning, prehending upstream task-speciﬁc vectors to steer a downstream model. In this way, we could only store a small amount of parameters to achieve model custom on downstream tasks. Furthermore, Lester~\cite{lester-etal-2021-power} used the special token embedding as soft prompt initialization, which need not prehend parameters in front of each layer. Qin~\cite{qin-eisner-2021-learning} proposed to learn a mixture of soft prompts conditioned on fixed FM. Liu~\cite{liu2021gpt} proposed to construct inputs with anchor words and trainable vectors. They proved task-related anchor words could bring further improvement. PTR~\cite{HAN2022182} composed prompts with manual templates and virtual tokens, then tuned with rules. Gu~\cite{gu-etal-2022-ppt} applied the pre-training on soft prompt to obtain a better initialization. Liu~\cite{liu2023pre} provide a detail survey of white-box tuning development. Jiang~\cite{jiang2023rethinking} rethought the past classic model tuning methods include prefix tuning, prompt tuning and adapter tuning, then unified them in a parallel form called U-Tuning. U-Tuning can incorporate off-the-shelf tuning methods and derive a framework for parameter efficient tuning. Wen~\cite{wen2023hard} applied the gradient-based discrete prompt optimization on multi-modal generation models, which can automatically design hard prompt for CLIP model.

\textbf{Black-Box Prompt Tuning:} Currently, considerably powerful large pre-trained models are deployed on the cloud, such as ChatGPT and GPT-3~\cite{brown2020language}. The users have no access to the parameters and gradients of models. This black-box setting secures the model owner from potential attack and misuse. As for commercial consideration, black-box service will become mainstream. However, querying cloud service through hand-crafted prompts cannot fully exploit data in many use cases. The urgent works mainly focus on black-box prompt tuning.

To this end, Sun~\cite{sun2022black} invokes derivative-free optimization on continuous prompt tuning. Concretely, they use a projection matrix to optimize the subspace of the original prompt.
Then prompts are sampled and updated from the multivariate normal distribution, called BBT.
And later, Sun~\cite{sun-etal-2022-bbtv2} further proposed BBTV2, constructing a divide-and-conquer method to prepend and optimize layer-specific prompts.
Furthermore, Diao~\cite{diao2022black} design a policy gradient inspired framework characterizing the problem as discrete tokens selection problem. Specifically, they conduct a variance-reduced policy gradient algorithm to estimate the gradient of the categorical distribution. However, user needs to upload their to fine-tuned the cloud model, which damage the privacy. Xiao~\cite{xiao2023offsite} formulated an emulator simulating cloud model and an adapter for fine-tuning downstream tasks. And model owner only transfer the emulator and adapter to user for user-end fine-tuning, called off-site tuning.
Hou~\cite{hou2022promptboosting} paired prompts with corresponding output distribution into a large of weak learners. Then they use a gradient-free method, ADABOOST algorithm, to refine a pool of prompts.
RLPROMPT~\cite{deng2022rlprompt} optimized the discrete prompts with reinforcement learning, and the reward comes from black-box models.




\subsubsection{Instruction Tuning}
Prior works have shown that instruction tuning, the technique of fine-tuning FM on a set of NLP tasks formatted with instructions, further enhances the ability of the model to perform invisible tasks from instructions~\cite{liu2023visual,longpre2023flan,peng2023instruction}. Instruction tuning refers to the process of deep training of FM on a data set consisting of (instruction, output) pairs in a supervised manner. 
Different from prompt tuning, if prompt tuning is to guide the model to generate relevant content through prompts, then instruction tuning is to train the model to perform specific tasks through instructions.

FLAN~\cite{wei2021finetuned} takes a pre-trained language model with 137B parameters and adjusts the instructions on more than 60 NLP datasets expressed through natural language instruction templates and evaluates them on invisible task types. T0~\cite{sanh2021multitask} increases the number of tasks and prompts based on FLAN, proving that implicit multi-task learning can improve model generalization and zero-shot ability. InstructGPT~\cite{ouyang2022training} uses a reinforcement learning (RLHF) technique via human feedback to rank multiple outputs of the model based on the results of user and API interactions, and then uses this data to fine-tune GPT-3 so that the InstructGPT model is better at following instructions than GPT-3. Peng~\cite{peng2023instruction} used GPT-4 to generate command follow data for LLM tuning, and experiments showed that the 52K English and Chinese command follow data generated by GPT-4 had better zero fire performance on the new mission than the command follow data generated by the previous state-of-the-art model. Jang~\cite{jang2023exploring} in multi-task hinted fine-tuning found that the distributed approach of training a separate expert LM for each training task has many advantages, including (1) avoiding the negative task transfer that often occurs during instruction tuning, (2) being able to continuously learn new tasks without having to retrain previous tasks to avoid catastrophic forgetting, and (3) demonstrate combinatorial capabilities when merging individual experts.

\begin{figure}
  \centering
  \includegraphics[width=1\columnwidth]{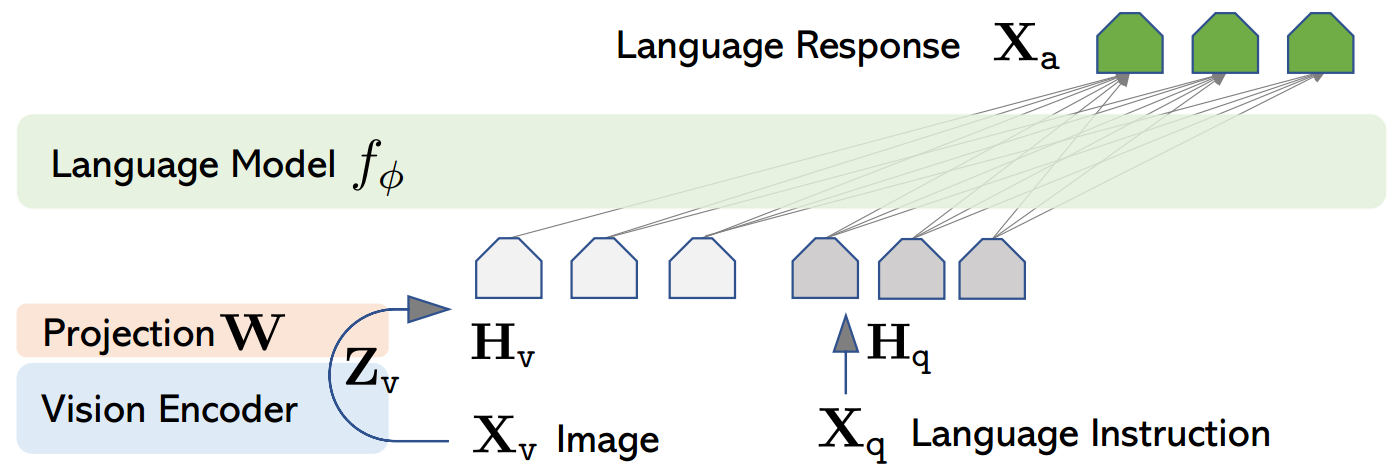}
  \caption{LLaVA network architecture~\cite{liu2023visual}}
  \label{fig: LLaVA}
\end{figure}

MULTIINSTRUCT~\cite{xu2023multiinstruct} datasets provide the basis for multimodal instruction tuning. Liu~\cite{liu2023visual} explored the design of instruction tuning in the field of multimodality, where they first attempted to use language-only GPT-4 to generate multimodal language image instruction follow data. Instruction tuning of this generated data enabled end-to-end training of a large multimodal model, LLaVA, capable of connecting visual encoders and LLMS for general visual and language understanding. Similar work is being studied in the~\cite{li2023otter}. LLaVA network architecture is shown in Figure~\ref{fig: LLaVA}.

The improvements still needed for instruction tuning are (1) increasing the number of instruction tasks and model size to improve the performance and zero-shot capabilities of the model on new tasks. (2) A more precise instruction template is required, and attention should be paid to the difference between training and inference instruction template settings. (3) Pay attention to the balance of data under different tasks, and the number of training samples under each task does not need to be too large. 

\subsection{Database Augmentation}
In the practical application of FM (ChatGPT, LLaMA), the strategy of using external knowledge bases for database augmentation to further improve the effectiveness is gaining more and more recognition~\cite{borgeaud2022improving,yasunaga2022deep,wang2023shall}. Database augmentation improves the performance of FM in a different way than the model editing described in Section~\ref{sec:model editing}, especially if the internal knowledge of the model is insufficient and lagging. The model based on database augmentation allows the introduction of external knowledge from different sources including training corpus, external data, unsupervised data and other options. The database augmentation model typically comprises a retriever and a generator. The retriever retrieves relevant knowledge from external sources based on the query, while the generator combines the query with the retrieved knowledge to make model predictions.

The benefits of this strategy are numerous: (1) It enables larger models to acquire richer knowledge, especially up-to-date information. Despite the vast knowledge capacity of the large model, it does not retain all of the knowledge, especially the new knowledge that emerges after its training. (2) Alleviating the illusion problem of FM. By providing external information, we can make the output of large models more reliable and closer to the real world. For example, when using language models to comment on events, embedding causal logic between events can make the output of the larger model more reasonable. (3) Many of today's open source FM are generic, and combined with domain-specific external knowledge bases, these models can perform better in dealing with domain-specific problems. This not only improves the efficiency of the model, but is also a low-cost strategy for applying large models. By applying retrieval augmentation strategies to FM, we can give these models a greater range of knowledge, better handle complex real-world problems, and provide more accurate and in-depth responses in specific domains.

\subsubsection{Language Database Augmentation}
As deep learning has proven successful, some retrieval systems have now adopted dense learned representations that are based on the activations of a neural network. Continuous cache~\cite{grave2016improving} assigns probability values to tokens that have similar previous and current activation vectors, thereby expanding the model's context to include local history. The kNN-LM~\cite{khandelwal2019generalization} proposed by Khandelwal in 2020, expands upon this idea and applies it to transformers. Furthermore, it broadens the retrieval database to encompass the entire English Wikipedia, which has led to significant improvements in the Wikitext103 evaluation.

\begin{figure}[ht]
  \centering
  \includegraphics[width=1\columnwidth]{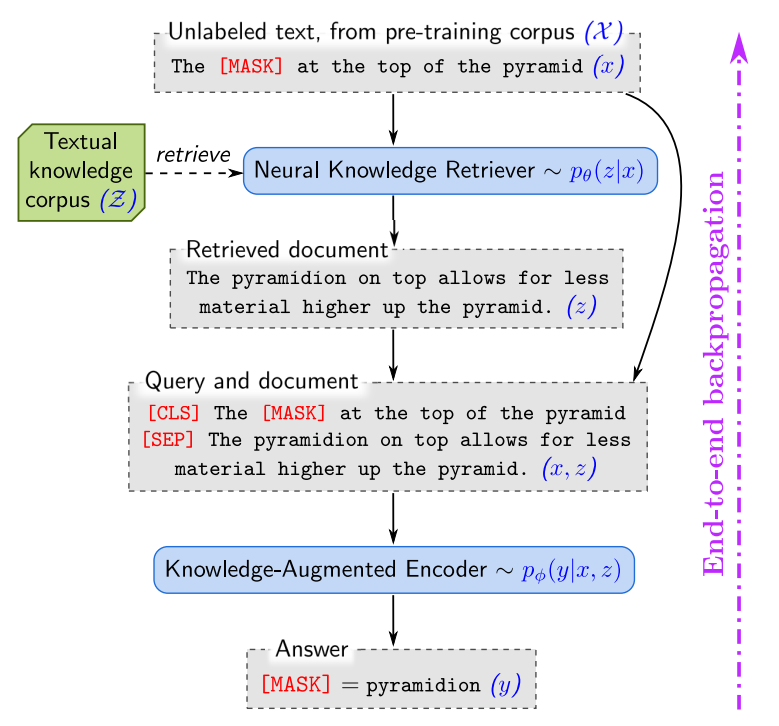}
  \caption{REALM augments language model pre-training with a neural knowledge retriever that retrieves knowledge from a textual knowledge corpus~\cite{guu2020retrieval}.}
  \label{fig:REALM}
\end{figure}

As shown in Figure~\ref{fig:REALM}, although the concept of search augmentation has been partially investigated before, REALM~\cite{guu2020retrieval} is the first to use "augmentation" as a whole concept for the optimization of pretrained models. By introducing a knowledge finder during the pre-training phase, the language model can explicitly use knowledge from text corpus (e.g., Wikipedia) during the pre-training, fine-tune, and prediction phase. 
With the introduction of REALM, retrieval augmentation is gradually emerging in the field of open-domain question answering (Open-QA). RAG~\cite{lewis2021retrievalaugmented} targets knowledge-intensive tasks, using pre-trained seq2seq models as parametric memory, and using pre-trained neural retrievers to access Wikipedia's dense vector index as non-parametric memory. RAG focuses on solving the shortcomings of previous studies that are limited to specific tasks. FID~\cite{izacard2020leveraging} focuses on how generative models handle retrieved paragraphs. This allows scaling to large numbers of documents and benefits from large amounts of evidence. Inspired by knowledge distillation technology, FID-KD~\cite{izacard2020distilling} considers the problem of annotated supervision of retrieved documents by using the attention score of reader model based on FID. Atlas~\cite{izacard2022few} works with the searcher to fine-tune the encoder-decoder model by modeling documents as latent variables. Similar to FID, RETRO~\cite{borgeaud2022improving} combines a frozen BERT searcher, a differentiable encoder, and a block-cross attention mechanism to merge retrieved text to predict tags with an order of magnitude more data than would normally be consumed during training. The difference is that RETRO performs retrieval in the pre-training phase, rather than insertion to solve a specific downstream task. REPLUG~\cite{shi2023replug} adds the retrieved documents to the input of the frozen black box large model, proving that it can be used to oversee the retrieval model, which can then find documents that help the large model make better predictions.

\subsubsection{Multimodal Database Augmentation}
Not limited to the plain text knowledge, database augmentation is also gaining attention in the field of multimodality, which means that searchers and generators need to be designed to process multimodal documents composed of images and text. MuRAG~\cite{chen2022murag} accesses external nonparametric multimodal memory to enhance language generation. Using joint contrast and generation loss, MuRAG was pre-trained using a hybrid corpus of large-scale image text and plain text to implement the first multimodal retrieval augmentation transformer. Re-Imagen~\cite{chen2022re} explores the retrieval augmentation of text image generation models, using retrieved information to generate high-fidelity and faithful images. Considering that the generators in the above work are limited to a single mode, either text generation or image generation, Yasunaga~\cite{yasunaga2023retrieval} proposes a retrieval model RA-CM3 which can retrieve and generate text and image simultaneously. The input data and external memory consists of a set of multimodal documents, each of which is an arbitrary sequence of text/images, providing a generic and modular retrieval augmentation framework for multimodal models.

\subsection{Discussion And Application}
The main purpose of model tuning is to generalize FM to a specified downstream task. Weight tuning is mainly oriented towards white-box scenarios, where the parameters and structure inside the model are known, allowing for model modification. Input engineering can deal with model generalization problems in both scenarios, and database augmentation enhance model performance by introducing external knowledge base. The common point of these methods is that the feedback given by the model is needed for parameter adjustment, and the model optimization is realized through LFM. Compared with the LFD paradigm that trains the model from scratch with sampled data, LFM requires less data and computing power, and can understand new tasks based on the existing general knowledge of the model.

In the future, we believe that the model tuning based on LFM paradigm can be studied in the following directions:

\begin{itemize}
  \item ~\textbf{Loss function design} New forms of model fine-tuning loss function need to be explored. For example, self-supervised learning or other types of unsupervised learning can be used to reduce the dependence of fine-tuning on downstream task data.
  \item ~\textbf{Balancing Memory Size and Retrieval Efficiency} With a large enough retrieval memory, the probability of retrieving an example that closely resembles the query becomes higher. Unfortunately, the downside of this is that the overall inference efficiency of the search-enhanced generative model is low due to the considerable retrieval overhead. Therefore, it is urgent to consider some methods to balance the size of the retrieval memory and the retrieval efficiency, such as the retrieval memory of data compression~\cite{li2022survey}.
  \item ~\textbf{Design of tailored retrieval metrics} Exploring the application of tailored metrics for retrieval could potentially enhance the control we have over text generation. Existing universal metrics, while generally effective, can inadvertently limit the diversity of retrieval results. By assembling a more varied set of retrieval outcomes, we can enhance the breadth of useful information covered. Therefore, the incorporation of multiple, distinct retrieval metrics could pave the way for higher quality generation in the future. This shift from a universal to a more customized approach in retrieval metrics could revolutionize the way we generate text.
\end{itemize}

%% file: section/distillation.tex
\section{Model Distillation}
\label{sec:model distillation}
\begin{figure*}
  \centering
  \includegraphics[width=\textwidth]{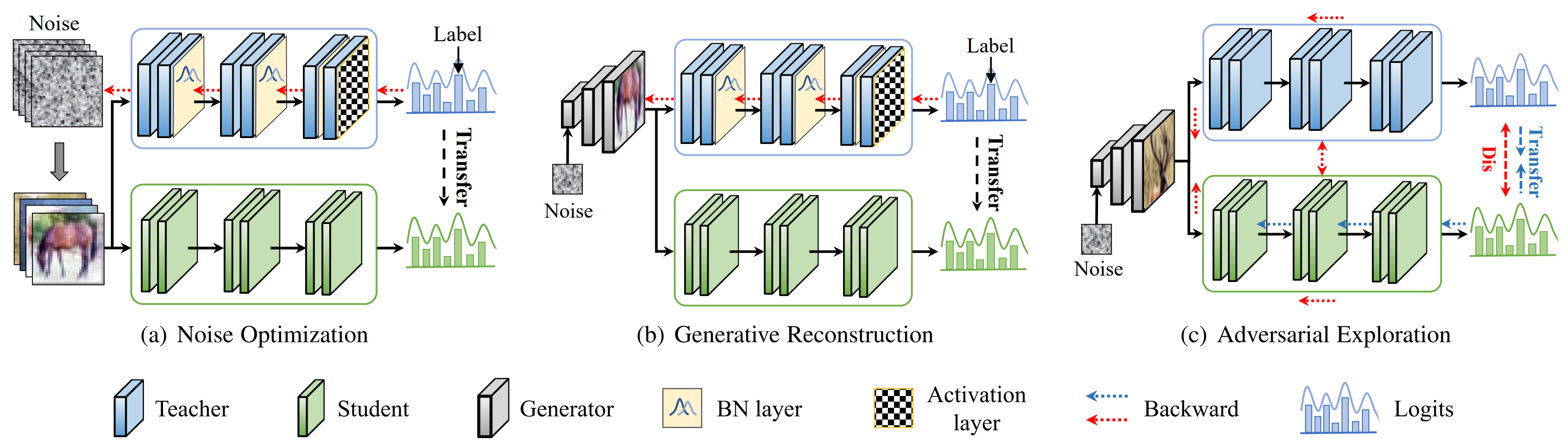}
  \caption{Three kinds of data-free knowledge distillation frameworks~\cite{liu2021data}.}
  \label{fig: KD}
\end{figure*}



Model Distillation is a knowledge distillation technique primarily used to compress deep learning models, reducing computational complexity, increasing runtime speed, and saving storage space while maintaining the performance of the original model~\cite{hinton_distilling_2015}. We show three traditional knowledge techniques in Figure~\ref{fig: KD}. The knowledge can be derived from various sources, such as the soft-label predictions~\cite{hinton_distilling_2015}, hidden layer activation~\cite{romero_fitnets_2015, koratana_lit_2019}, embedding~\cite{chen_learning_2018, ahn_variational_2019} or relationship among embeddings~\cite{ye_distilling_2020}.
With emergency of FM, researchers began to explore whether FM can be used as databases~\cite{petroni-etal-2019-language} to teacher student model without access to the training data.
The key insight in it is how to induce informative knowledge from FM.
In previous studies, these works were known as data-free knowledge distillation (DFKD)~\cite{liu_data-free_2021, braun_deep_2023, hu_learning_2023}.
In a white-box scenario, we possess access to the teacher model's architecture, gradient, parameters, and weights. Conversely, in a black-box scenario, we lack prior knowledge of the underlying model's architecture and parameters. Instead, each model is offered as a service (MaaS)~\cite{roman2009model}.
Building upon previous research, DFKD methods can be classified into three distinct paradigms: noise optimization, generative reconstruction, and adversarial exploration.

\subsection{Noise Optimization}
The core idea of noise optimization is leveraging teacher model to recover the alternative data. Specifically, these recovered data contain certain semantic information of training data~(e.g., intermediate feature map). We formulate the goal of noise optimization as following:
\begin{align}
  \hat{x} = \mathop{\arg\min}R(x|P)
\end{align}
where the $R$ is a regularization function combined with model $P$. Following the traditional knowledge distillation, the quality of recover data will affect the performance of student model, \ie, the $R$ is the key factors of whole process. And in this step, classifier loss or other auxiliary loss function can help with convergence.

The second step aims to align the prediction distributions of teacher and student models conditioned on recover data. The formulation of this step is below:
\begin{align}
  \mathop{\min} \sum_{\hat{x},\hat{y}} \mathcal{L}_{CE}(Q(\hat{x},\hat{y})) + \mathcal{L}_{KL}\left(\beta(Q(\hat{x})), \beta(P(\hat{x}))\right),
\end{align}
where the $\mathcal{L}_{CE}$ and $\mathcal{L}_{KL}$ indicate cross entropy loss and Kullback-Leibler (KL) divergence~\cite{joyce2011kullback}, $\beta(\cdot)$ represent the softmax function.

In terms of first step, many works endeavored to explore a better regularization function. Since the goal of distillation is alignment, main implementation method of this regularization function is also a distance function. Lopes~\cite{lopes2017data} proposed to minimize the distance of activation records~(\eg, means, covariance matrices). Concretely, this records come from teacher models' final layer before softmax. Furthermore, Dream Distillation~\cite{bhardwaj2019dream} use the statics of activation vectors as meta-data to generate images. Although, these methods claimed basing on source data-free setting. Indeed, we also need meta-data~(\ie, activation records) to optimize the noise. On the other hand, Nayak~\cite{nayak2019zero} proposed zero-shot Knowledge Distillation, synthesizing the Data Impressions from the complex Teacher model. Specifically, they utilize a Dirichlet distribution to model the softmax space of the Teacher network, where the concentration parameter of the Dirichlet distribution is interpreted as the measure of similarity among the components in the softmax vector. Therefore, they build a similarity matrix through the weight of pre-final layer and final layer before softmax function. Then this similarity matrix can be assigned as concentration parameter of Dirichlet distribution. Then based on "data impression", they can optimize the noise to synthetic pseudo data.

DeepDream~\cite{mordvintsev2015inceptionism} conduct an image prior to drive the generated away from unrealistic images with no discernible visual information:
\begin{align}
  R_{prior} = \alpha_{tv}R_{TV}(\hat{x}) + \alpha_{l_2}R_{l_2}(\hat{x}),
\end{align}
where $R_{TV}$ and $R_{l_2}$ penalize the total variance and $l_2$ norm of $\hat{x}$, respectively, with scaling factors $\alpha_{tv}$ and $\alpha_{l_2}$. Furthermore, DeepInversion~\cite{yin2020dreaming} extend image regularization $R_{prior}$ with a new feature distribution regularization term. DeepInversion proposed to minimize the distance between feature map statistics. Supposing that these features are all Gaussian-distributed, the feature distribution regularization term can be formulated as:

\begin{align}
  R_{BNS}(\hat{x}) = \sum||\hat{\mu_l}(\hat{x})-\mu_l||_2^2 + ||\hat{\sigma_l}(\hat{x})-\sigma_l||_2^2
\end{align}
where $\hat{\mu_l}(\hat{x})$ and $\hat{\sigma_l}(\hat{x})$ represent the estimated mean and variance of the feature maps, respectively, corresponding to the l-th neural layer. Average statistics stored in the widely-used BatchNorm (BN) layers are more than sufﬁcient. 


In terms of using publicly shared gradients, Zhu~\cite{NEURIPS2019_60a6c400} and ~\cite{geiping2020inverting,yin2021see} try to recover data from graidents. These gradients are derived from contemporary multi-node learning systems, wherein a key concept is to minimize the distance between the current node and the central node. The quality of synthetic data for noise optimization methods depends on the selection of knowledge priors, which usually requires a large amount of computational cost to update parameters for noise optimization.

\subsection{Generative Reconstruction}
This line of works is based on Generative Adversarial Networks (GANs)~\cite{NIPS2014_5ca3e9b1} to generate data representations or feature maps with pre-defined labels. Specifically, a noise vector $z$ is input into the generator $G$, which then maps $z$ to the desired data distribution. Aforementioned, prior information can be used to constraint the probability of generative model. The optimization function can be formulated as:

\begin{align}
  \min \mathbbm{E}_{z\sim p_z(z)}[R(G(z,\hat{y})|P) + \mathcal{L}(G(z,\hat{y}), \hat{y})]
\end{align}
where $p_z(z)$ is the distribution of $z$, $P$ is the pre-trained model.

Chen~\cite{chen2019data} proposed DAFL based on GAN framework, which regarded the pre-trained teacher networks as a ﬁxed discriminator. Then generator is trained to approximate images given fixed teacher network. In iteration of training GAN, compact student network can efficiently learn by utilizing generated images and the guidance of the teacher network. In addition to the cross-entropy loss function, they also proposed two other very useful regularization functions to constrain the generator. The first regularization function is applied to activation values. Since intermediate representations from real images tend to receive higher activation values than those from noise, the L1 norm of the intermediate representations can be used as one of the regularization terms. Furthermore, the real data can induce lower entropy of teacher model. Therefore, the regularization terms can be formulated as:
\begin{align}
  R = - ||f_P||_1 + - \frac{1}{N}\sum_i^{N}\hat{h}_i\log(\hat{h}_i)
\end{align}
where $f_p$ is the representation vetor from teacher model and $h_i$ is softmax vector from teacher model.

Ye~\cite{ye2020data} expand previous methods with group-stack GANs. Specifically, group-stack generators are used to collect data impression. Then all intermediate representations pass to corresponding discriminators with  knowledge amalgamated from previous teacher networks. This can be seen as deeper version of DAFL.
Yoo~\cite{NEURIPS2019_596f713f} proposed KegNet, which consisted of two trainable module. The learnable generator create alternative data then feed them into learnable decoder and fixed classifier. The generator created fake data conditioned on noise vector and pre-defined label. The decoder aimed to re-extract noise vector updating the errors of generator. The classifier was pre-trained teacher network, produced the label distribution of each fake data point. Finally, the generated images could boost knowledge distillation. DeGAN~\cite{addepalli2020degan} leveraged data from a related domain~(Proxy Data) to train GANs. Compared to typical GAN networks, DeGAN simply adds retrieval-enhanced samples to the training process of the discriminator. And following the privous work, DeGAN also adopted cross-entropy loss and a diversity loss conditioned on fixed classifier. As for proxy data, discriminator further penalized to fake data similar to them, which is a new class data.

Privious studies demonstrated that generated samples are similar in each class, which is interpreted as mode collapse issue in data reconstruction. Han~\cite{han2021robustness} introduce a new regularization function for improving the sample diversity within each class, called diversity seeking regularization. Concretely, they use $l_2$ distance to represent the similarity between the samples, which was added in loss function to pull away the generated samples. Chen~\cite{chen2021learning} filtered the large amount of wild data to further improve student networks. Firstly, they built a selection method to retrieve useful unlabeled images. Then the noisy label from teacher network are collected into a dataset. Under the noisy adaptation matrix supervised, portable student network was learned in conventional distillation manner and refined softmax space by noisy adaptation matrix.

The generation reconstruction method uses the generator to synthesize data, and reduces the training time by parameterizing the distribution of fake data in the iterative generation process. However, the quality of the synthesized data is seriously limited by the generator structure.

\subsection{Adversarial Exploration}
The main idea behind adversarial exploration based on adversarial generated strategies. The adversarial objective derived from on the discrepancy between the student and teacher networks, formulated as following:
\begin{align}
  D(P,Q;G),\label{eq:adv_object}
\end{align}
where the $P$ and $Q$ indicates pre-trained teacher model and random initialized student model, respectively; $G$ represents the generator.

As for generator updating, the generator creates images that continually widen the gap between the teacher and student networks, while the student and teacher networks are fixed in place, i.e, maximizing the distance in Eq.\ref{eq:adv_object}.
This objective is then used to update the generator.
Subsequently, we obtained these challenging samples that mislead the student model. In traditional distillation manner, student networks try to reduce the distance from teacher network conditioning on their predicted distributions or intermediate features. And the generator is fixed. 

Micaelli~\cite{Micaelli2019ZeroShotKT} proposed firstly a novel adversarial algorithm based on zero-shot knowledge transfer. They used KL divergence to search those samples poorly predicted by student network. Then traditional distillation is used to minimize KL divergence between teacher and student models. Based on previous methods, Fang~\cite{fang2019data} divided the generated images into hard and easy samples by estimating the true discrepancy between teacher and student models. Since hard samples always resulted in large output differences, an upper bound can be formulated for the real model discrepancy. Then they simply measures the Mean Absolute Error~(MAE) as the optimizing objective. Fang~\cite{fang2022up} proposed the FastDFKD framework, which achieved even 100× acceleration. Inspired by that in-domain instances always shared common features, these features can be reused to compose different sample without repetitive generating. In this way, FastDFKD built an efﬁcient synthesizer which captured common features for fast adaptation. He~\cite{he2023is} provide thorough experiments to investigate synthetic data to improve image recognition. 

Adversarial Exploration uses a truly adversarial framework for simultaneous knowledge extraction, the challenge being that the samples generated are not completely reliable, so some irrelevant samples are less effective in data transfer.

\subsection{Discussion And Application}
We have analyzed and reviewed three FM-based DFKD frameworks above, and in this section, we briefly discuss the application and future direction of DFKD methods.
\begin{itemize}
  \item \textbf{Integration with downstream methods:} Data-free knowledge distillation based on FM can be flexibly deployed in various excellent learning solutions, such as adversarial learning~\cite{patel2023learning}, automatic machine learning~\cite{rani2022machine}, meta-learning~\cite{lee2022meta}, and reinforcement learning~\cite{matsuo2022deep}. Therefore, combining knowledge distillation with other learning options will help solve the practical challenges of the future. 
  \item \textbf{Model applicability study:} The application conditions of model distillation are still harsh, the existing distillation algorithms are not stable enough, and most of the research is conducted on small data sets, and there is a lack of unified benchmark methods. In order to achieve perfect results, trial and error costs are high. This flaw is magnified infinitely in the face of the foundation model. We think it is also important to explore the influence of FM internal knowledge representation on the performance of model distillation and further design the distillation structure.
\end{itemize}

%% file: section/model_reuse.tex
\section{Model Reuse}
\label{sec:model reuse}

In the real-world training process, it is common practice to train multiple models with various hyperparameter settings. Subsequently, the best-performing model is selected or an ensemble of models is created to achieve superior performance \cite{dietterichEnsembleMethodsMachine2000}.
However, ensemble models require additional computing resources at inference time, and selecting a single model often results in discarding the remaining models.

In the real-world training process, it is crucial to experiment with different hyperparameter settings to explore the model's potential.
Training multiple models with diverse hyperparameter configurations allows us to capture a wider range of information and uncover optimal solutions.
However, the manual selection of the best model among these trained models can be a daunting task.
It often requires extensive computational resources, as training and evaluating each model can be time consuming.
Ensembling, on the other hand, offers a powerful solution to leverage the knowledge from multiple models.
By combining the outputs or predictions of individual models, we can create an ensemble model that exhibits improved performance.

\subsection{Ensemble Methods}

Ensemble learning is a widely used and effective approach in machine learning that involves combining predictions from multiple models to enhance overall performance. 
This paradigm unifies "weak" models by utilizing their collective outputs to create a more robust and accurate predictive model \cite{dietterichEnsembleMethodsMachine2000,sagi_ensemble_2018}.
By combining the predictions of multiple models, ensemble learning can help reduce the impact of biases and errors present in individual models and improve the generalization and stability of the resulting system.

Given a finite collection of models, represented as $\mathcal{M} = \{M_i\}_{i=1}^n$, where $n$ defines the total number of models in the set,  the ensemble of logits is defined as follows:
\begin{equation}
  M^* = \operatorname{Ensemble}(\mathcal{M}, \bm{\lambda}) = \sum_{i=1}^{n} \lambda_i M_i(\cdot),
\end{equation}
where $\lambda_i$ corresponds to a set of model-specific weight hyperparameters. These weight hyperparameters determine the relative importance of each model in the ensemble. The predictions of each model $M_i$ are weighted by the corresponding $\lambda_i$ and summed to generate the final prediction.
There has been significant research in this area, and has been applied in various domains, including computer vision \cite{DBLP:journals/corr/ZhouF17}, natural language processing \cite{aniol_ensemble_2019,wang_chinese_2016} and speech recognition \cite{deng_ensemble_nodate}.

\textbf{Efficient Ensemble}:
In \cite{wenBatchEnsembleAlternativeApproach2020}, the authors introduce BatchEnsemble whose computational and memory costs are significantly lower than typical ensembles. They accomplish this by using a shared weight for all ensemble members and combining it with a rank-one matrix specific to each member.
In \cite{liuTangentModelComposition2023}, the authors propose an ensemble method within the tangent space.
The method hinges on the equivalence of the ensemble of linearized models, i.e., the first-order Taylor approximations around the pre-trained weight, to a singular linearized model.
The primary strength of this framework lies in its ability to reduce the inference time of an ensemble of $n$ tangent models from $\mathcal{O}(n)$ to $\mathcal{O}(1)$.

\subsection{Model Fusion}

Ensemble methods such as bagging, boosting, or stacking provide effective techniques for combining models and can lead to enhanced generalization and robustness.
However, creating an ensemble involves running multiple models simultaneously, each model in the ensemble independently processes data and generates predictions, which are then aggregated to make a final decision, resulting in significant computational resources at inference time.

To address these challenges, this subsection aims to provide insights and strategies on how to effectively perform knowledge fusion.
By fusing the knowledge from multiple models, we can leverage the strengths of each model and enhance overall performance.
Advanced techniques for model fusion are being developed. These techniques aim to extract and combine the most useful information from each individual model to create a more powerful one \cite{li2023deep}.

\subsubsection{Weight Interpolation}

Weight interpolation is a simple yet effective method for model fusion.
This method involves combining the weights of multiple models to create a new model of the same network structure.
In this section, we will introduce the basic concepts of weight interpolation and its variants. We will also discuss some of the applications of weight interpolation.

Commonly, the subject of exploration is linear weight interpolation.
It is established knowledge that weights from distinct models, when fine-tuned on an identical dataset, can be associated via linear trajectories while maintaining a constant loss \cite{frankleLinearModeConnectivity2020,yunis2022on}. This concept is recognized as "linear mode connectivity".
The basic operation of linear weight interpolation is simple. For example, suppose that we have a model $M_1$ and a model $M_2$ that have been trained on the same task. Let $W_1$ and $W_2$ be the weights of $M_1$ and $M_2$, respectively. Then, by interpolating the weights of $M_1$ and $M_2$, we can create a new model $M_{new}$ that combines the knowledge from both $M_1$ and $M_2$. This new model $M_{new}$ can be defined as follows:
\begin{align}
  W_{new} = \lambda W_1 + (1-\lambda)W_2,
\end{align}
where $\lambda$ is a hyperparameter that controls the contribution of each model. This simple technique can be used to combine two models and create a new model that may exhibit enhanced performance.
Furthermore, this technique can be extended to combine multiple models. Suppose that we have $n$ models $M_1, M_2, ..., M_n$, then we can create a new model $M_{new}$ that combines the knowledge from all $n$ models by interpolating their weights as follows:
\begin{align}
  W_{new} = \lambda_1 W_1 + \lambda_2 W_2 + ... + \lambda_n W_n,
\end{align}
where $\lambda_i$ is a hyperparameter that controls the contribution of each model. This technique can be used to combine multiple models and create a new model that may exhibit enhanced performance.

\textbf{Uniform Averaging}:
Uniform averaging is the special case where $\forall_{i\in[n]}\lambda_i = 1/n$, and is also known as `model soups' or `adapter soups' for parameter-efficient fine-tuned models \cite{wortsman2022model,chronopoulou_adaptersoup_2023}.
The practice of uniform weight averaging is a straightforward but potent technique, which has found utilization in the domain of deep learning.
This method involves combining the weights of multiple models to create a new model with the same network structure, thereby raising the effective model reuse.
Its application helps improve generalization, as evidenced in~\cite{izmailovAveragingWeightsLeads2019}, while also providing superior accuracy by averaging the weights of models that are fine-tuned with diverse hyperparameter configurations \cite{wortsman2022model}.
Additionally, it increase robustness against distribution shifts~\cite{wortsmanRobustFinetuningZeroshot2022}.
Even just utilizing weight-space averaging of adapters trained on different domains, rather than full parameter space, can bolster out-of-distribution capability of model \cite{lu-etal-2022-improving}.
In addition to averaging on trained models, checkpoint averaging along the trajectory of a training run speeds up convergence \cite{kaddourStopWastingMy2022} and improves test generalization \cite{sanyalUnderstandingEffectivenessEarly2023}.
The Federated Learning community offers solutions such as FedAvg, a decentralized learning method, where models are collectively learned by averaging their weights \cite{mcmahanCommunicationEfficientLearningDeep2023}.

\textbf{Weighted Averaging}:
In addition to the same weight value, different models can have different weight values, and even different layers of different models can have their own weight values.
For model-specific weighted averaging, the model-specific scalar hyperparameters $\lambda_i, i\in [n]$ sum up to $1$, i.e. $\sum_{i=1}^{n} \lambda_i = 1$.
There has been significant research in this area, with recent studies including \cite{ilharco_patching_2022, matena_merging_2022,jolicoeur-martineau_population_2023}.
Checkpoint averaging methods are also model-specific average methods but along the training trajectory. 
For example, latest weight averaging (LAWA) \cite{kaddourStopWastingMy2022, sanyalUnderstandingEffectivenessEarly2023}.

\textbf{Task Arithmetic}:
Gabriel~\cite{DBLP:conf/iclr/IlharcoRWSHF23} edited large pre-trained model with task vectors, which can be formulated as
\begin{equation}
  W_{new} = W_{pre} + \sum_{i=1}^{n} \lambda_i \underbrace{\left(W_{ft}^{(i)} - W_{pre}\right)}_{\tau_i},
\end{equation}
where $W_{pre}$ is the parameters of a pre-trained model, $W_{ft}^{(i)}$ is the parameters of model fine-tuned on $i$-th downstream task, $\tau_i:=W_{ft}^{(i)} - W_{pre}$ is named as task vector, and $\lambda_i$ are not explicitly bounded.
They found that the interpolated model's performance on specific downstream target task can be manipulated to be better via task vector addition or reduced via task vector negation.

\subsubsection{Mode Connectivity Based Method}

Neural network loss landscapes refer to the relationship between the model's parameters and the corresponding loss function.
Understanding these landscapes is crucial for optimizing neural networks.

The idea of mode connectivity refers to the findings that different local minimal in model weight space might be connected by simple paths along which the model's training or validation performance is nearly constant \cite{garipov_loss_2018, frankleLinearModeConnectivity2020, yunis2022on}.
These paths could potentially be found using optimization methods.
Finding mode connectivity is a relatively new field of research and several methods have been proposed.

Especially, if finetuned from a same pre-trained model, the neural network often lies in a single loss basin in weight space.

Git Re-Basin is a technique for aligning and merging models trained with different random initializations or hyperparameters by permuting units to match a reference model and merging in weight space, creating a convex basin of functionally equivalent weights near the reference \cite{ainsworth_git_2023}.
ZipIt builds on model merging techniques by allowing partial zipping of models up to a specified layer to account for differences in models trained on disjoint tasks. It also introduces a general ``zip'' operation to merge weights within each model. These changes enable feasible merging of arbitrary models trained on completely different tasks \cite{george_stoica_zipit_2023}.

\subsubsection{Straightforward Optimization}

Straightforward optimization methods for model fusion involve simple and direct techniques to combine models.
These methods typically focus on finding the optimal weightings or parameters for the fused model by minimizing an objective function. These methods provide a straightforward way to combine models without requiring complex algorithms or architectures.

In \cite{singhModelFusionOptimal2023}, a layer-wise model fusion algorithm that utilizes optimal transport to (soft-) align neurons across the model before averaging their associated parameters is proposed.
Various techniques aim to minimize the distance (such as the $l_2$ norm distance) between the merged and individual models. 
Some methods offering closed-form solutions \cite{jin_dataless_2023} and others employing gradient-based optimization approaches \cite{lou_towards_2020, matena_merging_2022}.

\subsection{Discussion And Application}

In real-world training, it is common to train multiple models with different hyperparameters and select the best model or ensemble them.
However, ensembles require more inference resources and selecting a single model often discards useful knowledge.
Ensemble methods like bagging and boosting combine models to improve generalization and robustness. However, they require running multiple models, increasing computational costs. Efficient ensemble methods like BatchEnsemble reduce costs by using a shared weight plus a low rank matrix per member \cite{wenBatchEnsembleAlternativeApproach2020}.
Model fusion extracts and combines useful knowledge from individual models into a unified, improved model. This avoids ensemble computational costs.
Some researches on model reuse have been carried out in transfer learning~\cite{wu_pi-tuning_2023, yang_boosting_2023}, model compression~\cite{wei_ntk-approximating_2023,wenBatchEnsembleAlternativeApproach2020}, contunual learning~
\cite{liu_tangent_2023}, domain adaptation, personalization~\cite{chen_federated_2023}, efficient search~\cite{wortsman2022model} and multi-task leanring~\cite{george_stoica_zipit_2023}.
The key goals of model reuse methods include:
\begin{itemize}
  \item \textbf{Model complementarity research:} Improving overall predictive performance by combining complementary knowledge from different models.
  \item \textbf{Reduce computing costs:} Reducing computational costs compared to ensembling multiple models.
  \item \textbf{Knowledge representation extraction:} Extracting essential knowledge from multiple models into a distilled representation.
  \item \textbf{Robustness improvement:} Building robustness by merging diverse models into a unified whole.
  \item \textbf{Enable efficient model training and editing:} Accelerating model training by transfer and reuse of knowledge from pre-trained models.
\end{itemize}

%% file: section/meta_learning.tex
\section{Meta Learning}
\label{sec:meta-learning}
Meta-learning improves the performance of a model on a new task by extracting experience (usually data distribution) from multiple related tasks. The core idea of meta-learning is that learning becomes more efficient with increased experience, facilitated by the acquisition of inductive biases or knowledge that enhances future learning~\cite{schmidhuber1996simple,wang2021meta}. The abstract formula for meta-learning can be expressed as follows:

\begin{equation}
  \min _\omega \underset{\mathcal{T} \sim p(\mathcal{T})}{\mathbb{E}} \mathcal{L}(\mathcal{D} ; \omega)
\end{equation}
where $\omega$ represents meta-knowledge, $p(\mathcal{T})$ corresponds to the distribution of tasks, and $\mathcal{D}$ represents the dataset associated with each task. Our goal is to acquire a universal meta-knowledge that minimizes the loss for various tasks, striving for improved performance across different tasks.

This ‘learning-to-learn’~\cite{wortsman2019learning} method can
lead to a variety of benefits such as improved data and compute efficiency, and it is better aligned with human and animal learning, where learning strategies improve both on
a lifetime and evolutionary timescales~\cite{zhang2022progressive,chi2022metafscil,nguyen2022transformer}.


The generalized use of meta-learning can be traced back to the~\cite{chan1993experiments}, proposed by Philip as a general technique for fusing the results of multiple learning algorithms. Subsequently, meta-learning paradigm has been introduced into many fields and widely used. In the field of biology, Bengio~\cite{bengio1990learning} proposes an original approach to neural modeling based on meta-learning to find biologically credible synaptic learning rules, which enable networks to learn to perform difficult tasks. The learning rules involved can be optimized in the parameter learning rule space ~\cite{bengio1995search}. 
Thrun~\cite{thrun2012learning} provides the first clear definition and analysis of learn to learnin the field of algorithms, with significant practical implications for the field of machine learning and beyond. Vilalta~\cite{vilalta2002perspective} believes that the goal of meta-learning is to build an adaptive learner (that is, a learning algorithm that dynamically improves its bias through experience by accumulating meta-knowledge), and points out how to use meta-knowledge to improve the performance of learning algorithms key to progress in the field.

In essence, meta-learning is also based on the model itself to extend to downstream tasks, and is therefore one of the important paradigms of LFM. This chapter focuses on the research progress and classification of meta-learning related to pretrained large model transfer learning. 
In particular, meta-learning has the potential to mitigate many of the major shortcomings of contemporary deep learning, such as improving data efficiency, facilitating knowledge transfer, and unsupervised learning. Meta-learning has been shown to be useful in multi-tasking scenarios, for example in areas such as small lens image recognition, reinforcement learning (RL), hyperparameter optimization, and neural structure search (NAS).

From the perspective of the LFM paradigm, meta-learning can be further referred to as data-free meta-learning (DFML)~\cite{wang2022meta}, which achieves effective generalization to new tasks by meta-learning from a collection of pre-trained models without access to training data. According to the degree of understanding of the internal parameters and structure of the model, DFML is specifically divided into white-box data-free learning and black-box data-free learning. White-box data-free meta-learning focuses on exploiting the model based on the known underlying architecture and parameters of the model. Black-box data-free learning takes into account the generalization of tasks under the premise that the internal structure of the model is unknown. This method is more suitable for the background of modern industrial systems that only open large-scale model APIs for access, such as OpenAI's GPT-4~\cite{openai2023gpt4}, Google' Lamda~\cite{thoppilan2022lamda}, etc., which is one of the main problems solved by the LFM paradigm.

\subsection{White-box Data-free Meta-learning}
\begin{figure}
  \centering
  \includegraphics[width=0.65\columnwidth]{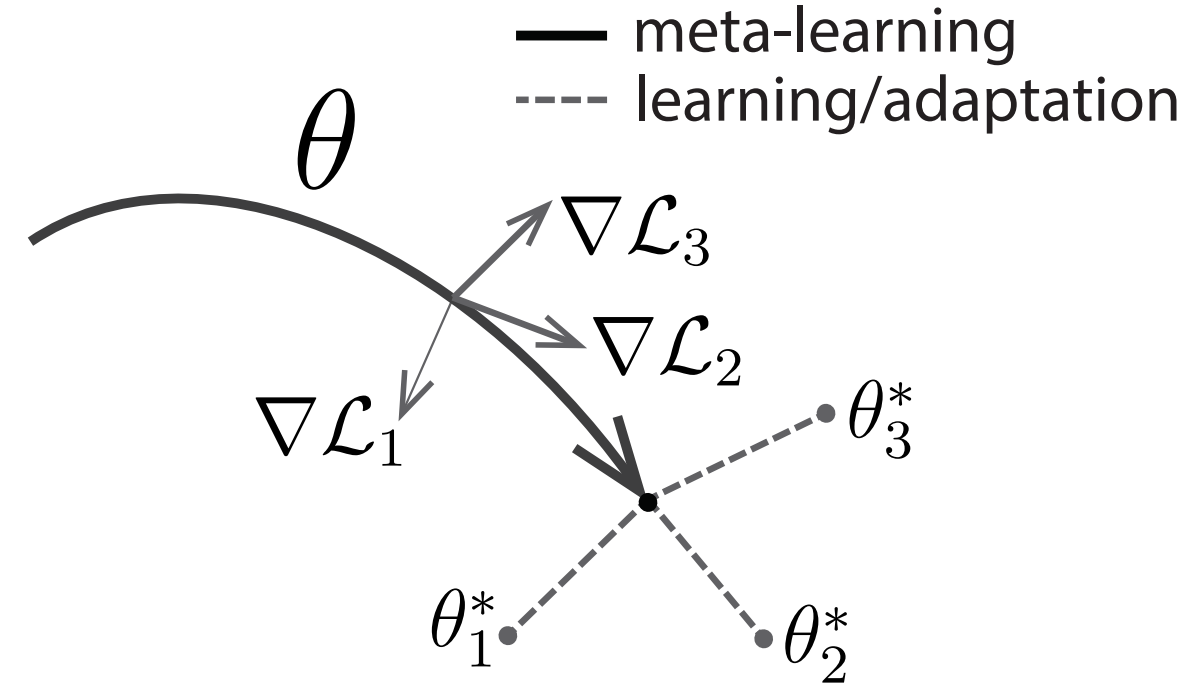}
  \caption{Schematic of the MAML algorithm, which optimizes a representation $\theta$ that can quickly adapt to new tasks~\cite{finn2017model}}
  \label{fig:MAML}
\end{figure}

As shown in Figure~\ref{fig:MAML}, MAML~\cite{finn2017model} is compatible with any model trained with gradient descent and can be applied to a variety of different learning problems, including classification, regression, and reinforcement learning, setting a precedent for optimization-based meta-learning. The key idea of MAML is to use a set of source tasks $\left\{T_1, \ldots, T_k\right\}$ to find the initialization of parameters $\theta_0$ from which learning a target task $T_0$ would require
only a small number of training samples.

Continual-MAML~\cite{caccia2020online} addresses the problem of fast online adaptation to new tasks while maintaining acquired knowledge on previously learned tasks. Since then, a large number of MAML variants require access to the original training data of the model~\cite{simon2022meta,wang20223meta,wang2022learning}, which restricts the application of meta-learning in large model migration. Existing white-box data-free learning methods solve this problem in parameter space. Meta-transfer learning (MTL)~\cite{sun2019meta} learning transfers large-scale pre-trained DNN weights to solve a small number of learning tasks, which opens the paradigm of white box learning. A similar technique is proposed in ~\cite{sun2020meta}, also for small sample learning. Meta-DETR~\cite{zhang2021meta} performs meta-learning object localization and classification at the image level in a small sample environment, using a semantic alignment mechanism (SAM) to encode and query supporting images into features of a specific class, and then input these features into a class-independent decoder to directly generate predictions for the specific class. CDCS~\cite{chai2022cross} adaptes a large common programming language pre-trained model CodeBERT~\cite{feng2020codebert} based on MAML to suit domain specific languages, and the experimental results are significantly better than the traditional pre-trained code model directly fine tuned.

\subsection{Black-box Data-free Meta-learning}
A lot of DFML work can only deal with pre-trained models in white-box scenarios, but this assumption is often difficult to meet, especially when the training cost of commercial large models is high, and meta-learning requires that downstream models also have the same architecture, limiting the scenarios that can be applied.

How to use the method of meta-learning to generalize the black box model for the downstream task is the primary consideration.
Wang~\cite{wang2022meta} proposed to predict the meta initialization through a meta-trained black-box neural network. For each task identity $T_i$, the task identifier is first embedded as the corresponding task embedding $e_i$. A black-box network is used to fuse different pre-trained models. Assuming the parameters of the black-box model are $\phi$, the black-box function $f_{\phi}(e_i)$ is utilized for fitting the parameters of task $T_i$ with task embedding $e_i$ as input. The objective is the optimization of the following:
\begin{equation}
  \max _\phi \sum_{i=1}^{i=N} \log P\left(\boldsymbol{\theta}_i \mid f_\phi\left(\boldsymbol{e}_i\right)\right).
\end{equation}

The likelihood function $P\left(\boldsymbol{\theta}i \mid f\phi\left(\boldsymbol{e}_i\right)\right)$, where $\sigma$ is a standard deviation constant, quantifies the deviation of the black-box network prediction from the true pre-trained task parameters.
\begin{equation}
  P\left(\boldsymbol{\theta}_i \mid f_\phi\left(\boldsymbol{e}_i\right)\right)=\exp \left(-\frac{\left\|f_\phi\left(\boldsymbol{e}_i\right)-\boldsymbol{\theta}_i\right\|^2}{\sigma^2}\right).
\end{equation}

Reptile~\cite{nichol2018first} extends MAML's results by repeatedly sampling a task for training and moving the initialization toward the task's training weights.
This approach still has some limitations, such as merging models only in the parameter space and ignoring the underlying data knowledge that can be extracted from the pre-trained model. At the same time, the way in which neural networks are used to predict model parameters determines that they can only be applied to small-scale pre-trained models. 
BiDf-MKD~\cite{hu2023learning} integrates model distillation and meta-learning, inverting each API to recover its training data only through a query API and a zero-step estimation, and then transferring more general meta-knowledge from a series of black-box apis to a single meta-model through a two-layer meta-knowledge distillation structure. PURER~\cite{hu2023architecture} further overcomes the limitations of data representation and model architecture on the basis of DFML and proposes a unified framework to synthesize a sequence of pseudo episodes.

In the field of computer vision, the small sample optimization method goes far with the help of meta-learning~\cite{ravi2016optimization}. RN~\cite{sung2018learning} learns to learn a deep distance metric to compare a small
number of images within episodes, each of which is designed to simulate the few-shot setting. FILM~\cite{jiang2023film} uses a pre-trained language model based on comparative learning, and adopts MAML to train the model through double-layer optimization to achieve model generalization with few samples.

Another important application of meta-learning in black box scenarios is in the field of model security (Meta Attack~\cite{du2019query}). Du~\cite{du2019query} proposes a meta-attack approach to learn generalizable prior abstractions from previously observed attack patterns to help infer attack patterns from a small number of queries and outputs. MGAA~\cite{yuan2021meta} improves the transferability of adversarial examples in black box Settings by narrowing the gap in gradient direction between white box attacks and black box attacks. In order to perform adductive attacks on black box models without allowing queries, Qin~\cite{qin2023training} designed a Meta-Surrogate Model (MSM), which is mathematically expressed as a two-layer optimization problem, and designed a differentiated attacker to make training feasible. This makes the adversarial examples generated on the MSM extremely transferable. The model extraction attack framework D-DAE~\cite{10179406} uses a meta-learning-based corruption detection module to learn the basic differences between the distribution of corrupted and uncorrupted query results.

\subsubsection{Discussion And Application}
Meta-learning can be seen as transfer learning in a two-tier optimization scenario, i.e. the embodiment of LFM. The study of meta-learning without data is of great significance to the field of large models, especially when the concept of model as a service is widely mentioned. The pre-trained model interface provided by various major manufacturers can not only be used as a tool to solve specific tasks, but also as a training resource for white/black box no-data learning to realize effective learning of new unseen tasks. The point of this is to eliminate the need to perform meta-learning on large amounts of labeled data and to reliably protect data privacy and security. At present, under the complementarity of FM and meta-learning, some researches have been carried out in few-shot image classification~\cite{zhu2022mgml}, robotic control policy learning~\cite{tang2023meta}, hyperparameter optimization~\cite{watanabe2023speeding}, meta-learning learning rules~\cite{stewart2022meta} and abstract reasoning~\cite{jiang2022role} have made excellent progress.

From the perspective of learning from models, We believe that there are still the following problems in DFML to be studied:
\begin{itemize}
  \item ~\textbf{Effective black box DFML algorithm} The primary focus of meta-learning research remains on the issue of data knowability, and studies on Data-Free Meta Learning (DFML) are still relatively limited. Given the current trend of pre-training model proliferation, the development of efficient and fast black-box DFML algorithms has become particularly crucial.
  \item ~\textbf{Design of general DFML algorithm} White-box DFML requires precise parameters for each pre-trained model and requires that all pre-trained models share the same architecture and cannot be adapted to large pre-trained models. Therefore, the research of general white box DFML algorithm is also one of the promising directions.
  \item ~\textbf{Cost optimization scheme} Meta-learning usually requires a highly complex optimization process, which makes DFML computationally complex and expensive. Especially in scenarios based on FM, the iteration of a large number of parameters will exponentially increase the computational cost. A considerable part of the work has put forward the cost optimization method of the meta-learning structure~\cite{zhang2022distributed,song2022efficient,gao2020modeling}, and the follow-up will be a direction worthy of efforts.
  \item ~\textbf{Zero-shot learning/Few-shot learning} In the few shot scenario, the performance of FM meta-learning is still weaker than the fine tuning paradigm, which restricts the landing deployment in the real scene~\cite{gama2023overview}. Improving the performance of meta-learning under small or even zero samples is also one of the research directions.
\end{itemize}

%% file: section/model_editing.tex
\section{Model Editing}
\label{sec:model editing}
Under the huge computational overhead of pre-trained models, updating the knowledge inside the model is not a simple "learning task". Ideally, with the complex transformation of various situations in the world, large models should also keep up with the pace of The Times anytime and anywhere~\cite{elazar2021measuring,lazaridou2021mind}, but the computational burden of training new large models cannot make large models achieve instant updates, so the algorithm for knowledge updating on the basis of the original pre-trained model has received widespread attention. Model editing, as one of the paradigms of LFM, can effectively change the knowledge of a pre-trained model in a specific domain without adversely affecting the results of other inputs. Model editing can better serve other LFM paradigms, because all LFM paradigms have high requirements for knowledge storage and output accuracy of the model itself.

To be clear, although both involve parameter and structure adjustments, model editing and model fine-tuning are not the same task oriented. The primary purpose of model fine-tuning is to use the pre-trained model for downstream tasks, while model editing focuses on updating the overall knowledge base or domain knowledge base within the model from a macro perspective, and does not need to be designed specifically for one task. Model editing can be considered a prerequisite for model fine-tuning.

\begin{figure}[ht]
  \centering
  \includegraphics[width=1\columnwidth]{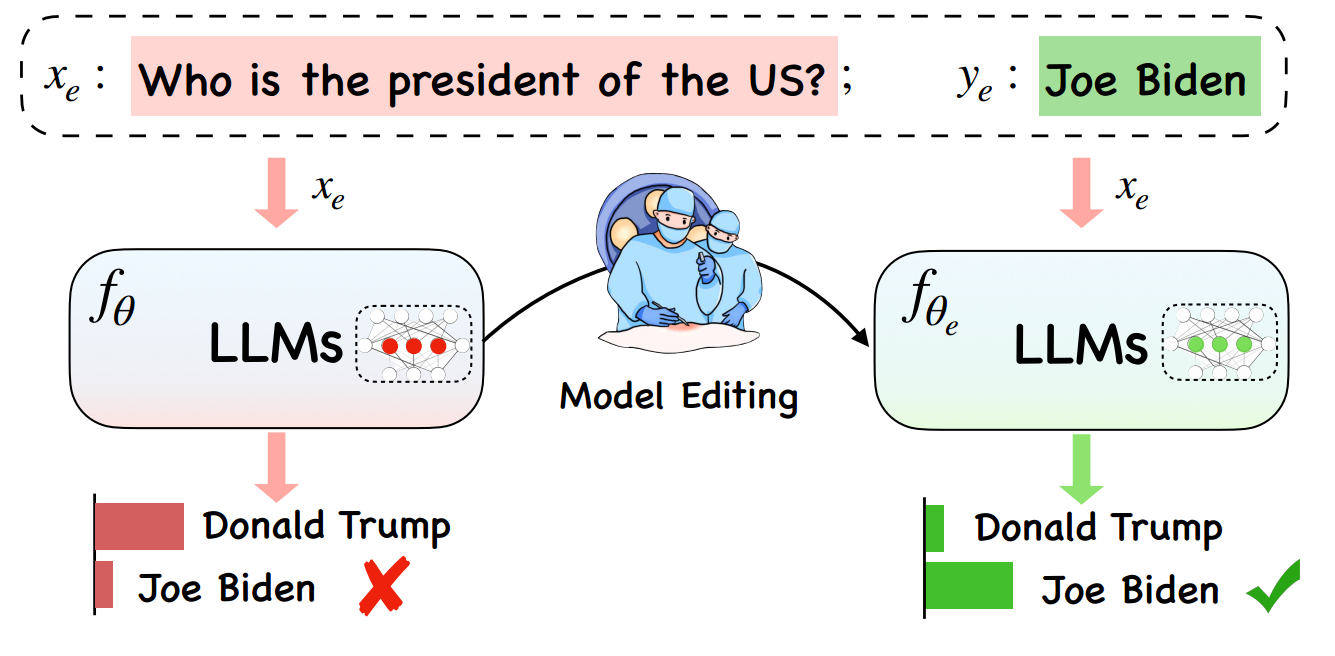}
  \caption{Model editing techniques for updating large pretrained models~\cite{yao2023editing}.}
  \label{fig:Model Editing}
\end{figure}
As shown in Figure~\ref{fig:Model Editing}, $f_{\theta}(x) \neq  y$, our intention is to adjust the original pretrained model $f_{\theta}$ by using the edit descriptor $(x_e,y_e)$ to get the modified model $f_{\theta_e}$, and $f_{\theta_e}(x) =y$. On the other hand, model editing also requires that editing in the current domain does not affect the normal output results of inputs in other domains~\cite{yao2023editing}, which can be formally expressed as:
\begin{equation}
  f_{\theta_e}(x_e)= \begin{cases}y_e & \text { if } x_e \in I_{x_e} \\ f_\theta(x_e) & \text { if } x_e \in O_{x_e}\end{cases}
\end{equation}
When the input $x_e$ is in the domain $I_{x_e}$ that needs to be updated, the output is adjusted to $y_e$, otherwise the original model output $f_\theta(x_e)$ remains.

The concept of model editing was first defined in 2020 ~\cite{sinitsin2020editable}, KnowledgeEditor~\cite{de2021editing} implements model editing on BERT architecture language models. Mitchell~\cite{mitchell2021fast} brings model editing to the field of large pre-trained models. According to the concrete implementation of current model editing in the field of large-scale pre-trained models, we mainly divide it into four categories: (1) adding additional parameters while freezing the original model parameters, and (2) directly modifying the internal parameters of the model. The first method is also called memory based model editing, and the core idea is to keep the original parameters of the model unchanged, by adding an additional set of independent parameters to modify the results of the model output. The second method acts as parameter based model editing, modifying the original model parameters directly to achieve the desired effect.

\subsection{Memory Based Model Editing}

Memory based model editing Uses additional editors (such as a new network) to influence the behavior of the model and keep the original parameters unchanged. An external learning editor was developed that modifies the fine-tuning gradient of the edits, but does not change the basic model that must process the edits ~\cite{hase2021language}.

SERAC~\cite{mitchell2021fast} wraps a black-box base model with an explicit cache of user-supplied edit descriptors (arbitrary discursions of the language model), as well as a small auxiliary range classifier and counterfactual model. Instead of making changes to the underlying model in the parameter space, SERAC simply stores the edits in the cache, using a range classifier to assess the likelihood that new input will fall within the scope of the stored edits. The drawback of SERAC as a learnable editor is that it relies excessively on editing datasets to train classifiers and counterfactual models. MemPrompt~\cite{madaan2023memoryassisted} introduces user feedback on PLMs error correction. Specifically, they keep a memory of errors in the model and user feedback, thereby enhancing the model to produce update prompts and avoid similar errors. Transformer-Patchcher~\cite{huang2023transformer} extends model editing to sequential model editing (SME), borrowing the idea of adapter tuning to change the behavior of Transformer-based models by simply adding and training a few neurons in the last feedforward network layer. CALINET~\cite{dong2022calibrating} also uses a specific FFN in the Calibrated FFN extended PLM for the addition of additional parameters, which consists of multiple calibration memory slots. These parameters are trained on the modified fact data set to achieve the effect of model editing. Model editing for black box scenarios is still worth exploring, and a recent article shows us the way~\cite{onoe2023can,murty2022fixing}.

\subsection{Parameter Based Model Editing}
An update matrix is applied to edit the model to update some internal parameters of the model~\cite{yao2023editing}. Parameter-based model editing can be finely divided into constrained tuning, locate and edit, and meta-learning.

\subsubsection{Constrained Tuning}
In order to combine new data and update the knowledge inside the model, a very natural and simple idea is to use new data for fine-tuning. In order to reduce the forgetting of knowledge, some restrictions need to be added in the process of fine-tuning, such as updating parameters as little as possible and only updating part of the structure of the model. Zhu~\cite{zhu2020modifying} formulates the model modification as a constraint optimization problem, constrines the loss of unmodified facts, and explores a better baseline method to approximate the execution of this constraint. In the implementation process, considering the maintenance cost of constraints, constraints are approximately expressed as:
\begin{equation}
  \text { minimize }_{\theta \in \Theta} \frac{1}{m} \sum_{x \in \mathcal{D}_{\mathcal{M}}} L(x ; \theta) \quad \text { subject to } \quad\left\|\theta-\theta_0\right\| \leq \delta \text {, }
\end{equation}
The original model is fine-tuned on the modified fact dataset $\mathcal{D}_M$, while using any appropriate norm explicitly constrained in the parameter space to minimize interference with the unmodified facts. This constraint is approximated by using the local continuity of the loss around $\theta_0$.

\subsubsection{Locate And Edit}
By introducing the concept of knowledge neuron, Dai~\cite{dai2021knowledge} makes a preliminary study on how fact knowledge is stored in pre-trained Transformers, and proposes a knowledge attribution method to identify neurons that express this fact. The activation of these knowledge neurons is positively correlated with the expression of the corresponding facts, and knowledge neurons can be directly utilized to edit (such as updating and deleting) specific factual knowledge without fine-tuning. MEMIT~\cite{meng2022mass} uses explicitly calculated parameter updates to insert new memories for weights of transformer modules identified as causal mediators of factual knowledge recall, which can be expanded to mass store thousands of memories. Ilharco~\cite{ilharco2022editing} proposes a new paradigm for guiding neural network model editing centered on task vectors. A task vector is constructed by subtracting the weight of the pre-trained model from the weight of the same model after fine-tuning the task. Task vectors can be modified and combined by arithmetic operations such as negative numbers and additions, and the resulting model's behavior is controlled accordingly.

\subsubsection{Meta Learning}
The use of meta-learning methods to modify model parameters is also a feasible direction. Meta-learning methods use hypernetworks to learn the weight matrices needed to edit large pre-trained models. KnowledgeEditor (KE)~\cite{de2021editing} trains a supernetwork with constraint optimization to modify a fact without affecting other knowledge, and then uses the trained supernetwork to predict weight updates at test time. MEND~\cite{meng2022locating} integrates a network of small gradient-decomposed model editors to make fast local edits to the behavior of a pre-trained model using a single desired input/output pair. MEND learns to transform gradients fine-tuned by standard, using low-rank decomposition of gradients to make the parameterization of such transformations easy to handle, and trained MEND can quickly apply new edits to pre-trained models. This method directly modifies the parameters of the model, which is good for editing a single knowledge or a small amount of knowledge, but easy to cause model parameter collapse when applied to a large amount of knowledge editing.

\subsection{Discussion And Application}
Model editing is necessary to learn and use models (LFM) more rationally, helping to keep them in sync with the real world in real time. Therefore, efficient, reliable and accurate algorithm design is essential. Many of the methods mentioned above still have great limitations waiting for further research. Model editing technology is currently mainly targeted at tasks such as neural machine translation~\cite{raunak2022rank}, text style transfer~\cite{luo2023prompt} and text-image generation~\cite{zhu2023conditional} based on FM.

\begin{itemize}
  \item  \textbf{Model editing object}
        Current model editing research focuses on the editing of factual knowledge, which is relatively easy to formalize and evaluate, but other types of knowledge included in editing models, such as preferences, require more sophisticated algorithm design. Developing a unified editing framework for all types of knowledge of the model is also a feasible direction. The realization of editing these contents requires a sufficient understanding of the knowledge representation of the pre-trained model, and some work can provide beneficial guidance for the development of these works~\cite{jiang2020can,cohen2023crawling}.

  \item \textbf{Black box editting} Previous model editing work was still based on the premise that the internal structure of the pre-trained model was known. Quite a few large pre-trained models show excellent performance (ChatGPT, GPT4), but can only be accessed through apis. How to edit the model in the black box environment and use it for downstream tasks still needs a lot of research.

  \item \textbf{Unified evaluation benchmark} Most knowledge editing research utilizes metrics such as the success rate of editing target knowledge, the invariance rate of predictions on unrelated knowledge (to assess generality), and the accuracy of paraphrasing target knowledge (to assess consistency). However, we find these assessments limited in their ability to comprehensively evaluate the knowledge editing capabilities of different strategies. For example, many evaluations only sample unrelated knowledge from the same distribution as the target knowledge~\cite{cao2023life}. Therefore, it is crucial to design comprehensive benchmark tests that can more effectively assess the capabilities of editing strategies.

\end{itemize}

%% file: section/challenges.tex
\section{CHALLENGES AND FUTURE PROSPECT}
\label{sec:challenges}
\subsection{Summary and Research Trends}

\begin{figure*}[ht]
  \centering
  \includegraphics[width=1\linewidth]{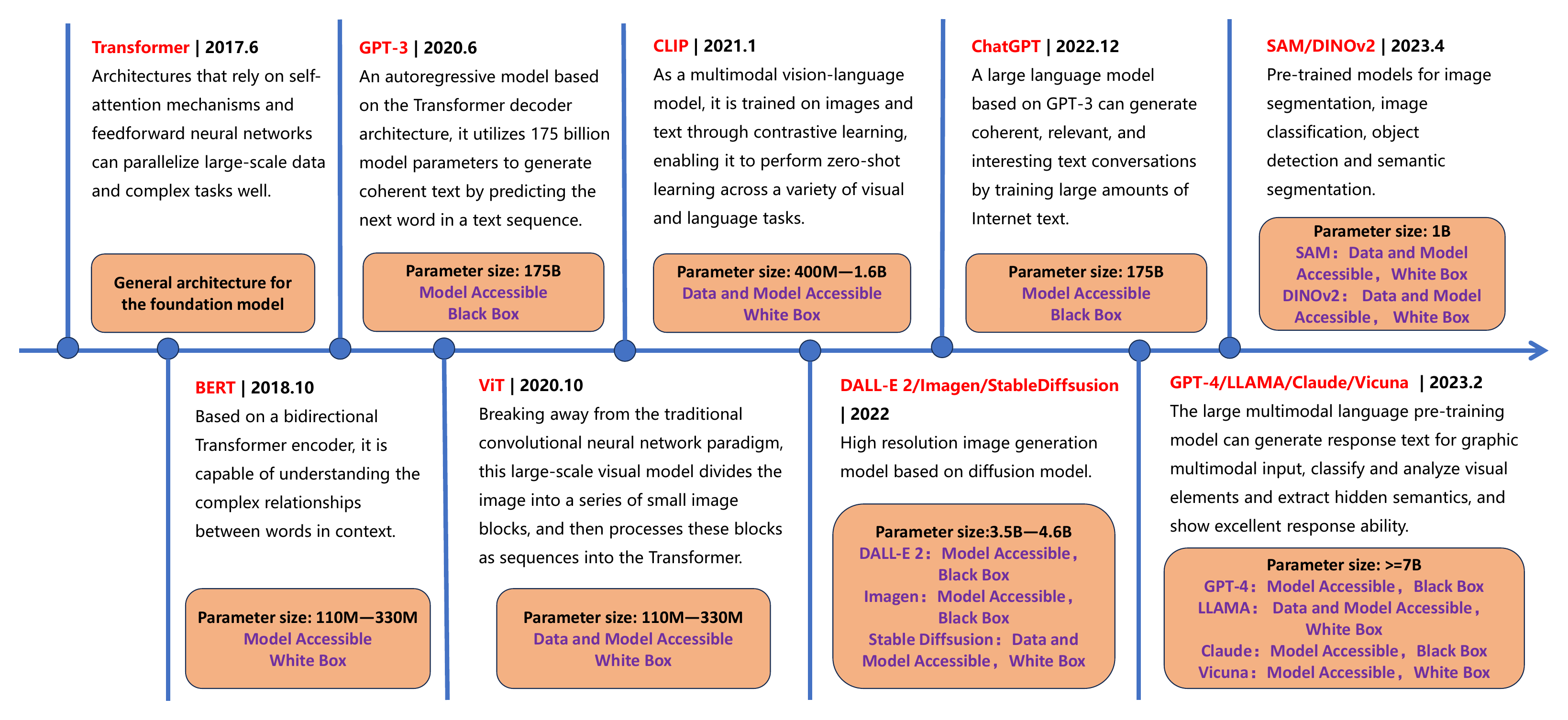}
  \caption{
    \textbf{The evolution of large language models and multimodal models}.
    This figure highlights the key milestones in the development of large pre-trained language models and multimodal models over the past several years. It traces the progression from bidirectional Transformer-based models like BERT to autoregressive models like GPT-3 and diffusion models for image generation like DALL-E 2.
  }
  \label{fig:time}
\end{figure*}

To provide a visual representation of the evolution of large language models, we have created a timeline in Figure~\ref{fig:time}. This timeline showcases the names and key contributions of prominent large models proposed by the academic community in recent years. By visually presenting this information, readers can gain a comprehensive understanding of the advancements and innovations taking place within the large-scale pre-training domain. Building upon the foundation laid by these models, researchers can further enhance their capabilities, explore new methods, and unlock new avenues to solve complex language understanding and generation challenges.

~\textbf{Interdisciplinary research on LFM} Interdisciplinary research is essential to advancing the science of machine learning, with various fields providing valuable insights and innovations through interdisciplinary research. The LFM paradigm itself provides specific and feasible generalization ideas for each discipline, and each discipline in turn provides the overall direction for LFM. In future work, we believe that actively promoting cooperation in different research fields and facilitating the exchange of ideas will be key to overcoming disciplinary barriers.

~\textbf{Multimodal large model Learning} Large language models exhibit surprising Zero/Few Shot reasoning performance on most NLP tasks, but they are inherently "blind" to vision because they can only understand discrete text. Large visual base models progress rapidly in perception and slowly in reasoning. Given this complementarity, the resulting multimodal architecture of language models and visual models running simultaneously toward each other is expected to make LFM more intelligent, including the advantages of having a more user-friendly interface, being a more comprehensive task solver, and better fitting the way humans perceive the world.

~\textbf{Model upgrade} With the supplement of data and the upgrade of algorithms, we believe that the perceptual pre-training model in the previous stage has completed the progress to the cognitive pre-training model, and the next step is to move towards the general pre-training model, that is, to use a more unified way to make the ability of the pre-training model land in the industry. From the perceptual intelligence of "listening, speaking, and seeing", to the cognitive intelligence of "thinking, answering questions, summarizing, translating, and creating", and even to the level of "decision-making and reasoning". The combination of pre-trained models and reinforcement learning is also a new trend.

~\textbf{Solving LFM problems with basic models} With the increasing complexity of parameters and structures of pre-trained models, the treatment of basic models shows great potential for solving large model problems in various research fields, making it an active and promising research area.

\subsection{Open Research Questions And Future Prospect}
In the following, we prospect several potential research
directions for future study.

~\textbf{Focus on the quality of training data}
Some of the biases in current models are actually caused by the training data itself. If the training data contains a disproportionate number of examples from a particular group, the model may learn to overpredict the outcome of that group. Conversely, if a certain group is underrepresented in the training data, the model may have difficulty accurately predicting the outcomes of members of that group. This uneven representation in the training data causes the model to be biased against certain outcomes, and this bias can perpetuate or even amplify existing social biases when the model is deployed. Ensuring the quality and representativeness of training data is critical. This includes carefully collating the data to ensure that it is comprehensive and representative of diverse groups and expected predicted outcomes. In addition, techniques such as data enhancement, rebalancing, and bias correction can be used to mitigate the effects of bias in training data.

~\textbf{In-depth analysis of knowledge representation inside large models} Models can be regarded as high-dimensional representations of data. With the increasing complexity of parameters and structures of pre-trained models, it has become an important research trend to explore how knowledge is represented inside models. Only when we fully understand the weight representation of the model and strengthen the research on the algorithm itself and the internal operation mechanism of the model can we save the cost better and apply the large-scale pre-training model to the downstream task more reasonably.

~\textbf{Addressing Security Concerns in LFM}
The strength of LFMs comes with its own set of security challenges, which are not dissimilar to those faced by smaller language models. These large, pre-trained models can be manipulated with unique instructions to generate harmful, biased, or offensive content that could be used maliciously, thereby posing a potential misuse risk. One of the primary methods to circumvent these issues is the application of reinforcement learning from human feedback (RLHF). This approach integrates human judgment into the training process, allowing for the creation of better-aligned models. Enhancing model safety can also be achieved by incorporating safety-centric guidelines during the RLHF phase. Despite its benefits, RLHF is heavily reliant on superior quality input from human annotators, making its practical implementation challenging. It is crucial, therefore, to refine the RLHF framework to lessen the burden on human annotators and explore more efficient annotation techniques to guarantee data quality. For instance, the use of models to facilitate annotation tasks could be a viable solution.

%% file: section/conclusion.tex
\section{Conclusion}
\label{sec:conclusion}

In this review, we provide a comprehensive review of the concepts of learning from models, covering methods such as model fine tuning, model distillation, model reuse, meta-learning, and model editing. Each approach uses a pre-trained model in a different way and in a different context for better performance or to solve a specific task. We systematically summarize relevant studies from early explorations to the present, not only providing a detailed overview of how each approach works, but also discussing their advantages and limitations. We took an in-depth look at how these approaches perform when dealing with real-world problems and the challenges that can be encountered during implementation. In addition, we also analyze the major challenges facing today, including issues such as interpretability, security, and fairness of models, and how to effectively fine-tune and edit models to adapt to new tasks or domains. Finally, we suggest some future directions, explore how these challenges can be addressed through further research, and how new technologies and approaches might advance the field. We believe this survey will benefit researchers in the fields of language models, knowledge graphs, knowledge repositories, and more, providing them with a comprehensive reference resource to understand current state-of-the-art methods and how they can be applied to improve their research and applications.